\documentclass{article}

\usepackage[nonatbib,final]{neurips_2022}

\usepackage{pythonhighlight}
\definecolor{keywordcolour}{RGB}{207,34,46}
\definecolor{stringcolour}{RGB}{26,62,115}
\definecolor{literatecolour}{RGB}{20,90,179}
\definecolor{specmethodcolour}{RGB}{137,90,225}
\usepackage{inconsolata}    %

\usepackage{dirtree}

\usepackage[utf8]{inputenc} %
\usepackage[T1]{fontenc}    %
\usepackage{hyperref}       %
\usepackage{url}            %
\usepackage{booktabs}       %
\usepackage{amsfonts}       %
\usepackage{nicefrac}       %
\usepackage{microtype}      %
\usepackage{pifont}
\usepackage{enumitem}
\usepackage{wrapfig}
\usepackage{siunitx}

\usepackage[backend=biber,natbib=true,style=authoryear,doi=false,isbn=false,url=false,eprint=false,giveninits=true,maxbibnames=200,maxcitenames=2,mincitenames=1,uniquename=false,uniquelist=false,dashed=false]{biblatex}
\usepackage[textwidth=3cm, textsize=scriptsize]{todonotes}

\definecolor{linkcolor}{RGB}{83,83,182}
\hypersetup{
    colorlinks=true,
    citecolor=linkcolor,
    linkcolor=linkcolor,
    urlcolor=linkcolor
}

\newcommand{\ie}{{\em i.e.,~}}

\newcommand{\eg}{{\em e.g.,~}}

\newcommand{\repo}[1]{#1}
\newcommand{\rebuttal}[1]{#1}

\usepackage{amsmath}
\newcommand{\norm}[1]{\left \Vert #1 \right \Vert}

\newcommand{\bbR}{\mathbb{R}}

\newcommand{\Benchopt}{{{\texttt{Benchopt}}}}
\newcommand{\Python}{{{\texttt{Python}}}}
\newcommand{\Julia}{{{\texttt{Julia}}}}
\newcommand{\PyTorch}{{{\texttt{PyTorch}}}}
\newcommand{\TensorFlow}{{{\texttt{TensorFlow}}}}
\newcommand{\Numba}{{{\texttt{numba}}}}
\newcommand{\Cython}{{{\texttt{Cython}}}}
\newcommand{\BLAS}{{{\texttt{BLAS}}}}
\newcommand{\Keras}{\texttt{Keras}}
\newcommand{\OpenML}{\texttt{OpenML}}
\newcommand{\MLPerf}{\texttt{MLPerf}}
\newcommand{\CSV}{\texttt{CSV}}
\newcommand{\HTML}{\texttt{HTML}}
\newcommand{\PDF}{\texttt{PDF}}
\newcommand{\skglm}{\texttt{skglm}}
\newcommand{\noncvxpro}{\texttt{noncvx-pro}}
\newcommand{\glmnet}{\texttt{glmnet}}
\newcommand{\cuML}{{{\texttt{cuML}}}}
\newcommand{\celer}{{{\texttt{celer}}}}
\newcommand{\blitz}{{{\texttt{blitz}}}}
\newcommand{\snapML}{{{\texttt{snapML}}}}
\newcommand{\lassojl}{{{\texttt{lasso.jl}}}}
\newcommand{\liblinear}{{{\texttt{liblinear}}}}

\newcommand{\argmin}{\mathop{\mathrm{arg\,min}}}
\DeclareMathOperator{\dist}{dist}
\usepackage[capitalize]{cleveref}

\newcommand{\myparagraph}[1]{\vspace{1mm}\noindent\textbf{#1} \,}

\usepackage{aliascnt}
\newaliascnt{problem}{equation}
\aliascntresetthe{problem}
\makeatletter

\def\endproblem{\eqno \hbox{\@eqnnum}$$\@ignoretrue}
\makeatother

\def\equationautorefname~#1\null{(#1)\null}
\def\problemautorefname~#1\null{Problem (#1)\null}

\title{Benchopt: Reproducible, efficient and collaborative optimization benchmarks}

\author{
  Thomas~Moreau$^{1,*}$,
  Mathurin~Massias$^{2,*}$,
  Alexandre~Gramfort$^{1,*}$,
  Pierre~Ablin$^{3}$,\\
  \textbf{
  Pierre-Antoine~Bannier, %
  Benjamin~Charlier$^{4}$,
  Mathieu~Dagréou$^{1}$,
  Tom~Dupré~la~Tour$^{6}$,}\\
  \textbf{
  Ghislain~Durif$^{4}$,
  Cassio~F.~Dantas$^{7}$,
  Quentin~Klopfenstein$^{8}$,
  Johan~Larsson$^{9}$,
  En~Lai$^{1}$,}\\
  \textbf{
  Tanguy~Lefort$^{4}$,
  Benoit~Malézieux$^{1}$,
  Badr~Moufad$^{2}$,
  Binh~T.~Nguyen$^{10}$,
  Alain~Rakotomamonjy$^{11}$,}\\
  \textbf{
  Zaccharie~Ramzi$^{12}$,
  Joseph~Salmon$^{4,5}$,
  Samuel~Vaiter$^{13}$}\\[1em]
  $^1$ Université Paris-Saclay, Inria, CEA, 91120 Palaiseau, France\\
  $^2$ Univ Lyon, Inria, CNRS, ENS de Lyon, UCB Lyon 1, LIP UMR 5668, F-69342, Lyon, France\\
  $^3$ Université Paris-Dauphine, PSL University, CNRS, 75016, Paris, France\\
  $^4$ IMAG, Univ Montpellier, CNRS, Montpellier, France\quad
  $^{5}$ Institut Universitaire de France (IUF)\\
  \hskip-7ex
  $^6$ University of California, Berkeley, CA 94720, USA\quad
  $^{7}$ TETIS, Univ Montpellier, INRAE, Montpellier, France\\
  $^8$ University of Luxembourg, LCSB, Esch-sur-Alzette, Luxembourg\\
  $^9$ The Department of Statistics, Lund University\quad
  $^{10}$ LTCI, Télécom Paris, 91120 Palaiseau, France\\
  $^{11}$ Criteo AI Lab, Paris, France\quad
  $^{12}$ ENS Ulm, CNRS, UMR 8553, Paris, France\\
  $^{13}$ CNRS \& Université Côte d'Azur, Laboratoire J.A. Dieudonné, CNRS, Nice, France\\
}

\newlength{\figwidth}
\begin{document}

\maketitle

\vspace{-20pt}
\begin{abstract}
    Numerical validation is at the core of machine learning research as it allows to assess the actual impact of new methods, and to confirm the agreement between theory and practice.
    Yet, the rapid development of the field poses several challenges: researchers are confronted with a profusion of methods to compare, limited transparency and consensus on best practices, as well as tedious re-implementation work.
    As a result, validation is often very partial, which can lead to wrong conclusions that slow down the progress of research.
    We propose \Benchopt{}, a collaborative framework to automate, reproduce and publish optimization benchmarks in machine learning across programming languages and hardware architectures.
    \Benchopt{} simplifies benchmarking for the community by providing an off-the-shelf tool for running, sharing and extending experiments.
    To demonstrate its broad usability, we showcase benchmarks on three standard learning tasks: $\ell_2$-regularized logistic regression, Lasso, and ResNet18 training for image classification.
    These benchmarks highlight key practical findings that give a more nuanced view of the state-of-the-art for these problems, showing that for practical evaluation, the devil is in the details.
    We hope that \Benchopt{} will foster collaborative work in the community hence improving the reproducibility of research findings.
\end{abstract}

\setlength{\figwidth}{\linewidth}

\section{Introduction}

\looseness=-1  %
Numerical experiments have become an essential part of statistics and machine learning (ML).
It is now commonly accepted that every new method needs to be validated through comparisons with existing approaches on standard problems.
Such validation provides insight into the method's benefits and limitations and thus adds depth to the results.
While research aims at advancing knowledge and not just improving the state of the art, experiments ensure that results are reliable and support theoretical claims \citep{sculley2018winner}.
Practical validation also helps the ever-increasing number of ML users in applied sciences to choose the right method for their task.
Performing rigorous and extensive experiments is, however, time-consuming \citep{raff2019step}, particularly because comparisons against existing methods in new settings often requires reimplementing baseline methods from the literature.
In addition, ingredients necessary for a proper reimplementation may be missing, such as important algorithmic details, hyperparameter choices, and preprocessing steps \citep{pineau2019iclr}.

In the past years, the ML community has actively sought to overcome this ``reproducibility crisis'' \citep{hutson2018reproducibility} through collaborative initiatives such as open datasets (\OpenML{}, \citealt{OpenML2013}), standardized code sharing \citep{forde2018reproducible}, benchmarks (\MLPerf{}, \citealt{mattson2020mlperf}), the NeurIPS and ICLR reproducibility challenges \citep{pineau2019iclr,pineau2021improving} and new journals (\eg{} \citealt{rougier2018rescience}).
As useful as these endeavors may be, they do not, however, fully address the problems in optimization for ML since, in this area, there are no clear community guidelines on how to perform, share, and publish benchmarks.

Optimization algorithms pervade almost every area of ML, from empirical risk minimization, variational inference to reinforcement learning \citep{sra2012optimization}.
It is thus crucial to know which methods to use depending on the task and setting \citep{bartz2020benchmarking}.
While some papers in optimization for ML provide extensive validations \citep{Lueckmann2021}, many others fall short in this regard due to lack of time and resources, and in turn feature results that are hard to reproduce by other researchers. In addition, both performance and hardware evolve over time, which eventually makes static benchmarks obsolete.
An illustration of this is the recent work by \citet{Schmidt2021}, which extensively evaluates the performances of 15 optimizers across 8 deep-learning tasks.
While their benchmark gives an overall assessment of the considered solvers, this assessment is bound to become out-of-date if it is not updated with new solvers and new architectures.
Moreover, the benchmark does not reproduce state-of-the-art results on the different datasets, potentially indicating that the considered architectures and optimizers could be improved.

{\hskip1em\parbox{.95\textwidth}{We firmly believe that this critical task of \textbf{maintaining an up-to-date benchmark in a field cannot be solved without a collective effort}. We want to empower the community to take up this challenge and \textbf{build a living, reproducible and standardized state of the art that can serve as a foundation for future research.}}}

\begin{figure}[t]
    \begin{center}
        \includegraphics[width=0.99\linewidth]{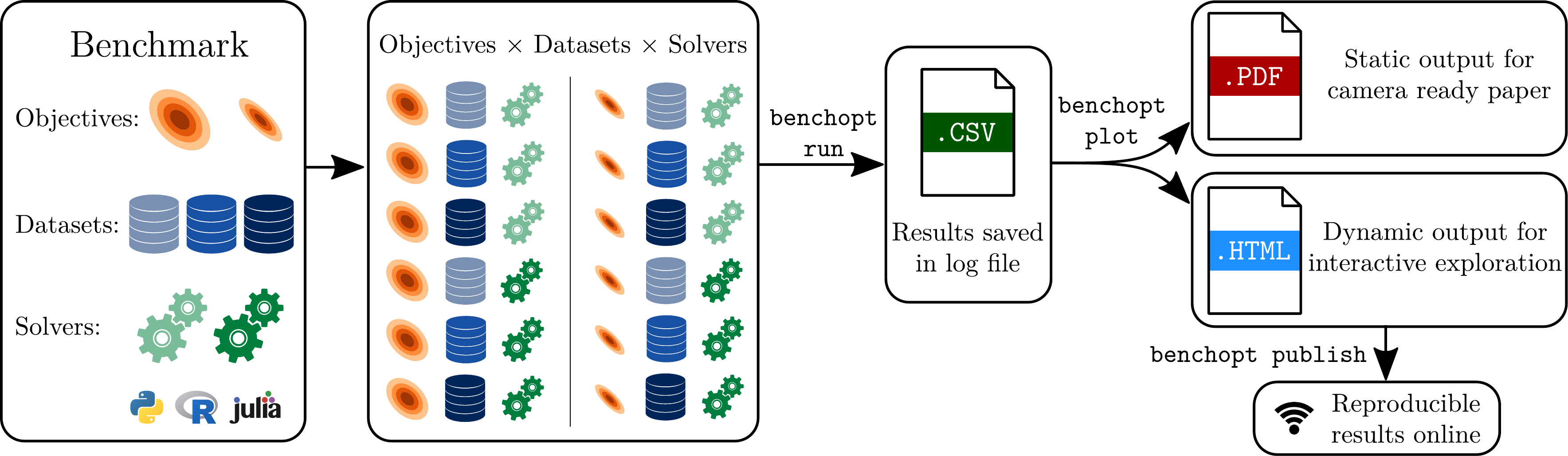}
        \caption{A visual summary of \Benchopt{}.
        Each \texttt{Solver} is run (in parallel) on each \texttt{Dataset} and each variant of the \texttt{Objective}.
        Results are exported as a \CSV{} file that is easily shared and can be automatically plotted as interactive \HTML{} visualizations or \PDF{} figures.}
        \label{fig:benchopt_visual_summary}
        \vspace{-20pt}
    \end{center}
\end{figure}

\Benchopt{} provides the tools to structure the optimization for machine learning (Opt-ML) community around standardized benchmarks, and to aggregate individual efforts for reproducibility and results sharing.
\Benchopt{} can handle algorithms written in \Python, \texttt{R}, \Julia{} or \texttt{C/C++} via binaries.
It offers built-in functionalities to ease the execution of benchmarks: parallel runs, caching, and automatical results archiving.
Benchmarks are meant to evolve over time, which is why \Benchopt{} offers a modular structure through which a benchmark can be easily extended with new objective functions, datasets, and solvers by the addition of a single file of code.

The paper is organized as follows.
We first detail the design and usage of \Benchopt{}, before presenting results on three canonical problems:
\begin{itemize}[topsep=0pt,itemsep=1ex,partopsep=0ex,parsep=0ex,leftmargin=3ex]
    \item $\ell_2$-regularized logistic regression: a convex and smooth problem which is central to the evaluation of many algorithms in the Opt-ML community, and remains of high relevance for practitioners;
    \item the Lasso: the prototypical example of non-smooth convex problem in ML;
    \item training of ResNet18 architecture for image classification: a large scale non-convex deep learning problem central in the field of computer vision.
\end{itemize}

The reported benchmarks, involving dozens of implementations and datasets, shed light on the current state-of-the-art solvers for each problem, across various settings, highlighting that the best algorithm largely depends on the dataset properties (\eg size, sparsity), the hyperparameters, as well as hardware.
A variety of other benchmarks (\eg MCP, TV1D, etc.) are also presented in Appendix, with the goal to facilitate contributions from the community.

By the open source and collaborative design of \Benchopt{} (BSD 3-clause license), we aim to open the way towards community-endorsed and peer-reviewed benchmarks that will improve the tracking of progress in optimization for ML.
\section{The Benchopt library}
The \Benchopt{} library aims to provide a \rebuttal{standard toolset and structure to implement}  benchmarks for optimization in ML, where the problems depend on some input dataset $\mathcal D$.
The considered problems are of the form
\begin{equation}\label{eq:benchopt_pb}
    \theta^* \in \argmin_{\theta\in\Theta} f(\theta; \mathcal D, \Lambda)\enspace,
\end{equation}
where $f$ is the objective function, $\Lambda$ are its hyperparameters,
and $\Theta$ is the feasible set for $\theta$.
The \textbf{scope} of the library is to evaluate optimization methods in their wide sense by considering the sequence $\{\theta^t\}_t$ produced to approximate $\theta^*$.
\rebuttal{We emphasize than \Benchopt{} does not provide a fixed set of benchmarks, but a framework to create, extend and share benchmarks on any problem of the form \eqref{eq:benchopt_pb}.}
To provide a flexible and extendable \rebuttal{coding standard}, benchmarks are defined as the association of three types of object classes:
\;\;\;\;\;\;\;\;\;\;\;\;\;\;\;\;\;\;\;\;\;\;\;\;\;\;
\;\;\;\;\;\;\;\;\;\;\;\;\;\;\;\;\;\;\;\;\;\;\;\;\;\;
\;\;\;\;\;\;\;\;\;\;\;\;\;\;\;\;\;\;\;\;\;\;\;\;\;\;
\begin{wrapfigure}{r}{3.1cm}
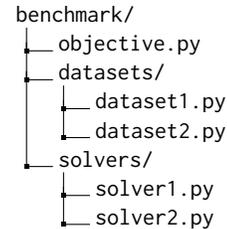

    \small%
    \dirtree{%
        .1 benchmark/.
        .2 objective.py.
        .2 datasets/.
        .3 dataset1.py.
        .3 dataset2.py.
        .2 solvers/.
        .3 solver1.py.
        .3 solver2.py.
    }
    \caption{Standard benchmark structure}
\end{wrapfigure}

\vspace*{-3mm}%
\begin{itemize}[topsep=0pt,itemsep=1ex,partopsep=0ex,parsep=0ex,leftmargin=2ex]
    \item[] \textbf{Objective:}
    It defines the function $f$ to be minimized
    as well as the hyperparameters $\Lambda$ or the set $\Theta$,
    and the metrics to track along the iterations (\eg objective value, gradient norm for smooth problems, or validation loss).
    Multiple metrics can be registered for each $\theta^t$.

    \item[]\textbf{Datasets:}
    The \texttt{Dataset} objects provide the data $\mathcal D$ to be passed to the \texttt{Objective} class.
    They control how data is loaded and preprocessed.
    \texttt{Datasets} are separated from the \texttt{Objective}, making it easy to add new ones, provided they are coherent with the \texttt{Objective}.

    \item[]\textbf{Solvers:}
    The \texttt{Solver} objects define how to run the algorithm.
    They are provided with the \texttt{Objective} and \texttt{Dataset} objects and output a sequence $\{\theta^t\}_t$.
    This sequence can be obtained using a single run of the method, or with multiple runs in case the method only returns its final iterate.
\end{itemize}

Each of these objects can have parameters that change their behavior, \eg{} the regularization parameters for the \texttt{Objective}, the choice of preprocessing for the \texttt{Datasets}, or the step size for the \texttt{Solvers}.
By exposing these parameters in the different objects, \Benchopt{} can evaluate their influence on the benchmark results.
The \Benchopt{} library defines an application programming interface (API) for each of these concepts and provides a command line interface (CLI) to make them work together.
A benchmark is defined as a folder that contains an \texttt{Objective} as well as subfolders containing the \texttt{Solvers} and \texttt{Datasets}.
\rebuttal{
\autoref{app:sec:code} presents a concrete example on Ridge regression of how to construct a \Benchopt{} benchmark while additional design design choices of \Benchopt{} are discussed in \autoref{app:design_choices}.
}

For each \texttt{Dataset} and \texttt{Solver}, and for each set of parameters, \Benchopt{} retrieves a sequence $\{\theta^t\}_t$ and evaluates
the metrics defined in the \texttt{Objective} for each $\theta^t$.
To ensure fair evaluation, the computation of these metrics is done off-line.
The metrics are gathered in a \texttt{CSV} file that can be used to display the benchmark results, either locally or
as \texttt{HTML} files published on a website that reference the benchmarks run with \Benchopt{}.
This workflow is described in \autoref{fig:benchopt_visual_summary}.

This modular and standardized organization for benchmarks empowers the optimization community by making numerical experiments easily reproducible, shareable, flexible and extendable.
The benchmark can be shared as a git repository or a folder containing the different definitions for the \texttt{Objective}, \texttt{Datasets} and \texttt{Solvers} and it can be run with the \Benchopt{} CLI, hence becoming a convenient reference for future comparisons.
This ensures fair evaluation of baselines in follow-up experiments, as implementations validated by the community are available.
\rebuttal{
Moreover, benchmarks can be extended easily as one can add a \texttt{Dataset} or a \texttt{Solver} to the comparison by adding a single file.
Finally, by supporting multiple metrics -- \eg{} training and testing losses, error on parameter estimates, sparsity of the estimate -- the \texttt{Objective} class offers the flexibility to define the concurrent evaluation, which can be extended to track extra metrics on a per benchmark basis, depending on the problem at hand.
}

\rebuttal{
As one of the goal of \Benchopt{} is to make benchmarks as simple as possible, it also provides a set of features to make them easy to develop and run.}
\Benchopt{} is written in \Python{}, but \texttt{Solvers} run with implementations in different languages (\eg \texttt{R} and \texttt{Julia}, as in \autoref{sec:lasso}) and frameworks (\eg \PyTorch{} and \TensorFlow{}, as in \autoref{sec:resnet18}).
Moreover, benchmarks can be run in parallel with checkpointing of the results, enabling large scale evaluations on many CPU or GPU nodes.
\rebuttal{
\Benchopt{} also makes it possible to run solvers with many different hyperparameters' values , allowing to assess their sensitivity on the method performance.
}
Benchmark results are also automatically exported as interactive visualizations, helping with the exploration of the many different settings.

\myparagraph{Benchmarks} All presented benchmarks are run on 10 cores of an Intel Xeon Gold 6248 CPUs @ 2.50GHz and NVIDIA V100 GPUs (16GB). \repo{The results' interactive plots and data are available at \url{https://benchopt.github.io/results/preprint_results.html}.}

\section{First example: \texorpdfstring{$\ell_2$}{l2}-regularized logistic regression}\label{sec:logreg}
Logistic regression is a very popular method for binary classification. %
From a design matrix $X \in \mathbb{R}^{n \times p}$ with rows $X_i$ and a  vector of labels $y \in \{-1, 1\}^n$ with corresponding element $y_i$, $\ell_2$-regularized logistic regression provides a generalized linear model indexed by $\theta^* \in \mathbb R^p$ to discriminate the classes by solving
\begin{problem}
    \label{pb:logreg}
  \theta^*
  = \argmin_{\theta \in \bbR^p}
    \sum_{i=1}^{n} \log\big(1 + \exp(-y_i X_i^\top \theta)\big) + \frac{\lambda}{2} \|\theta\|_2^2
    \enspace ,
\end{problem}
where $\lambda > 0 $ is the regularization hyperparameter.
Thanks to the regularization part, \autoref{pb:logreg} is strongly convex with a Lipschitz gradient, and thus its solution can be estimated efficiently using many iterative optimization schemes.

The most classical methods to solve this problem take inspiration from Newton's method~\citep{Wright1999}.
On the one hand, quasi-Newton methods aim at approximating the Hessian of the cost function with cheap to compute operators. Among these methods, L-BFGS~\citep{Liu1989} stands out for its small memory footprint, its robustness and fast convergence in a variety of settings.
On the other hand, truncated Newton methods \citep{Dembo1982} try to directly approximate Newton's direction by using \eg{} the conjugate gradient method~\citep{Fletcher1964} and Hessian-vector products to solve the associated linear system.
Yet, these methods suffer when  $n$ is large: each iteration requires a pass on the whole dataset.

In this context, methods based on stochastic estimates of the gradient have become standard \citep{DBLP:conf/compstat/Bottou10}, with Stochastic Gradient Descent (SGD) as a main instance.
The core idea is to use cheap and noisy estimates of the gradient~\citep{robbins1951stochastic,kiefer1952stochastic}.
While SGD generally converges either slowly due to decreasing step sizes, or to a neighborhood of the solution for constant step sizes, variance-reduced adaptations such as SAG \citep{Schmidt2017}, SAGA \citep{Defazio2014} and SVRG \citep{Johnson2013} make it possible to solve the problem more efficiently and are often considered to be state-of-the-art for large scale problems.

Finally, methods based on coordinate descent \citep{Bertsekas99} have also been proposed to solve \autoref{pb:logreg}.
While these methods are usually less popular, they can be efficient in the context of sparse datasets, where only few samples have non-zero values for a given feature, or when accelerated on distributed systems or GPU \citep{Dunner_18}.

\begin{figure}[t]
    \centering
    \includegraphics[width=.8\figwidth]{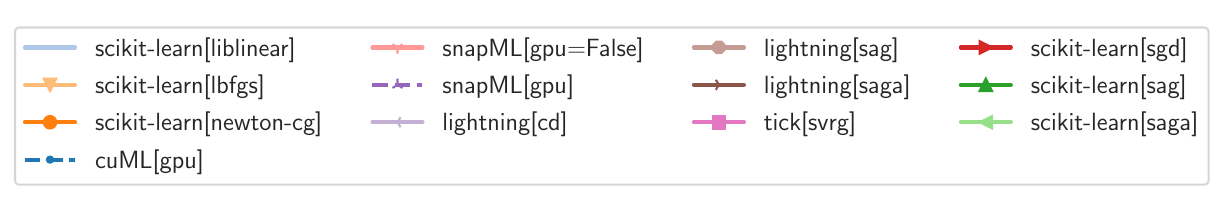}
    \includegraphics[width=0.99\figwidth]{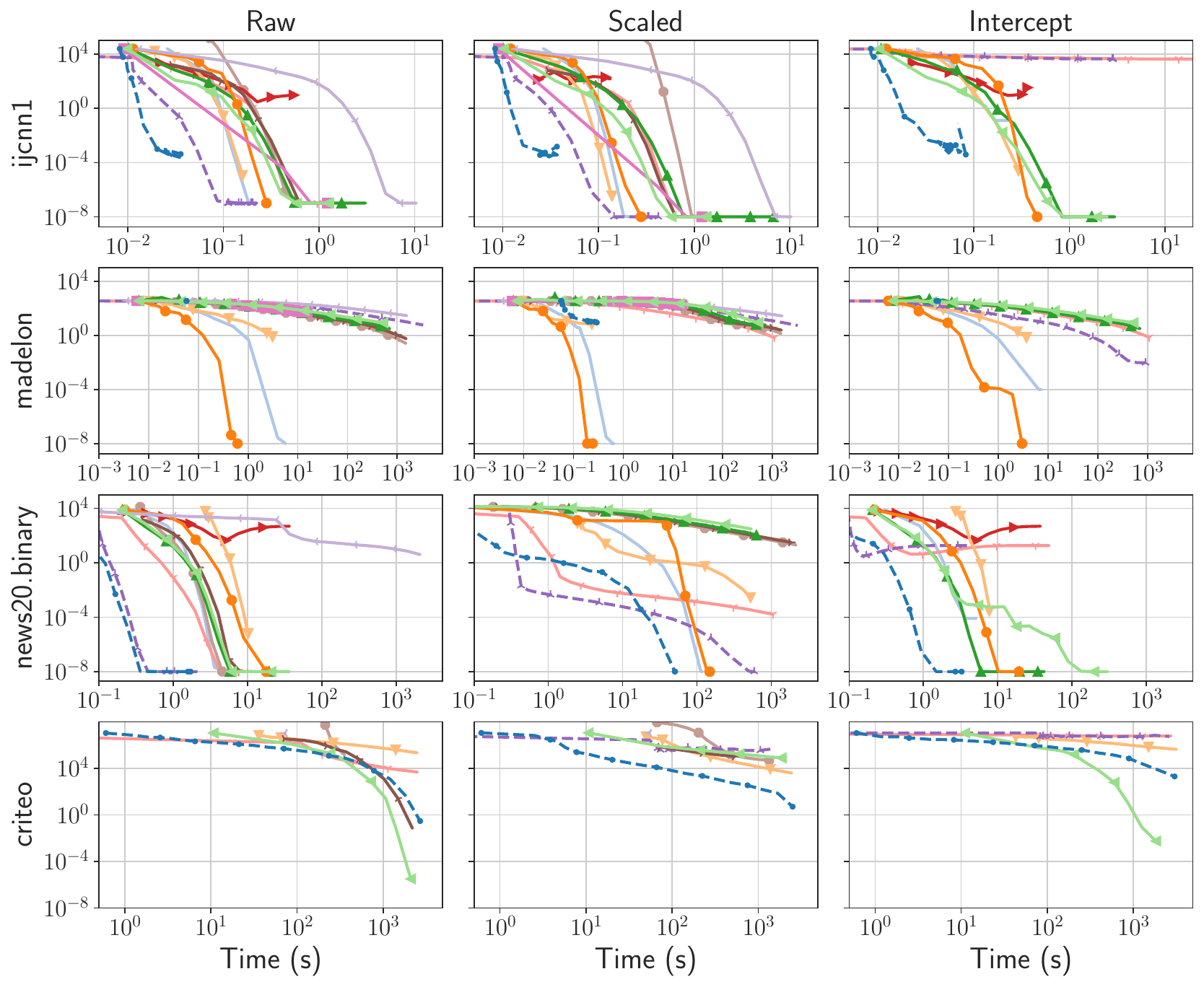}
    \caption{
        Benchmark for the $\ell_2$-regularized logistic regression, on 13 solvers, 4 datasets (\emph{rows}), and 3 variants of the \texttt{Objective} (\emph{columns}) with $\lambda = 1$.
        The curves display the suboptimality of the iterates, $f(\theta^t) - f(\theta^*)$, as a function of time.
        The first column corresponds to the objective function detailed in \autoref{pb:logreg}.
        In the second column, datasets were preprocessed by normalizing each feature to unit standard deviation. The third column is for an objective function which includes an unregularized intercept.
    }
    \label{fig:logreg_l2_medium}\vskip-1em
\end{figure}

\repo{The code for the benchmark is available at \url{https://github.com/benchopt/benchmark_logreg_l2/}.}
To reflect the diversity of solvers available, we showcase a \Benchopt{} benchmark with 3 datasets, 10 optimization strategies implemented in 5 packages, leveraging GPU hardware when possible.
We also consider different scenarios for the objective function:
    (i) \textbf{scaling} (or not) the features, a recommended data preprocessing step, crucial in practice to have comparable regularization strength on all variables;
    (ii) fitting (or not) an unregularized \textbf{intercept} term, important in practice and making optimization harder when omitted from the regularization term~\citep{Koh_Kim_Boyd07};
    (iii) working (or not) with \textbf{sparse} features, which prevent explicit centering during preprocessing to keep memory usage limited.
Details on packages, datasets and additional scenarios are available in \autoref{app:logregl2}.

\looseness-1
\myparagraph{Results}
\autoref{fig:logreg_l2_medium} presents the results of the benchmarks, in terms of suboptimality of the iterates, $f(\theta^t) - f(\theta^*)$, for three datasets and three scenarios.
Here, because the problem is convex, $\theta^*$ is approximated by the best iterate across all runs (see \autoref{app:sec:best_iterate}). %
Overall, the benchmark shows the benefit of using GPU solvers (\cuML{} and \snapML{}), even for small scale tasks such as \emph{ijcnn1}.
Note that these two accelerated solvers converge to a higher suboptimality level compared to other solvers, due to operating with 32-bit float precision.
Another observation is that data scaling can drastically change the picture.
In the case of \emph{madelon}, most solvers have a hard time converging for the scaled data. For the solvers that converge, we note that the convergence time is one order of magnitude smaller with the scaled dataset compared to the raw one. This stems from the fact that in this case, the scaling improves the conditioning of the dataset.\footnote{The condition number of the dataset is divided by~5.9 after scaling.}
For \emph{news20.binary}, the stochastic solvers such as SAG and SAGA have degraded performances on scaled data. Here, the scaling makes the problem harder.\footnote{The condition number is multiplied by~407 after scaling.}

On CPU, quasi-Newton solvers are often the most efficient ones, and provide a reasonable choice in most situations.
For large scale \emph{news20.binary}, stochastic solvers such as SAG, SAGA or SVRG --that are often considered as state of the art for such problem-- have worst performances for the presented datasets.
While this dataset is often used as a test bed for benchmarking new stochastic solvers, we fail to see an improvement over non-stochastic ones for this experimental setup.
In contrast, the last row in \autoref{fig:logreg_l2_medium} displays an experiment with the larger scale \emph{criteo} dataset, which demonstrates a regime where variance-reduced stochastic gradient methods outperform quasi-Newton methods. For future benchmarking of stochastic solvers, we therefore recommend using such a large dataset.

Finally, the third column in \autoref{fig:logreg_l2_medium} illustrates a classical problem when benchmarking different solvers: their specific (and incompatible) definition and resolution of the corresponding optimization problem.
Here, the objective function is modified to account for an intercept (bias) in the linear model.
In most situations, this intercept is not regularized when it is fitted. However, \snapML{} and \liblinear{} solvers do regularize it, leading to incomparable losses.
\section{Second example: The Lasso}
\label{sec:lasso}

The Lasso, \citep{Tibshirani96,Chen_Donoho_Saunders98}, is an archetype of non-smooth ML problems, whose impact on ML, statistics and signal processing in the last three decades has been considerable \citep{Buhlmann_vandeGeer11,Hastie_Tibshirani_Wainwright15}.
It consists of solving
\begin{problem}\label{pb:lasso}
    \theta^* \in \argmin_{\theta\in \bbR^p} \tfrac{1}{2} \norm{y - X\theta}^2 + \lambda \norm{\theta}_1 \enspace,
\end{problem}
where $X \in \mathbb{R}^{n \times p}$ is a design matrix containing $p$ features as columns, $y \in \mathbb{R}^n$ is the target vector, and $\lambda > 0$ is a regularization hyperparameter.
The Lasso estimator was popularized for variable selection: when $\lambda$ is high enough, many entries in $\theta^*$ are exactly equal to 0.
This leads to more interpretable models and reduces overfitting compared to the least-squares estimator.

Solvers for \autoref{pb:lasso} have evolved since its introduction by \citet{Tibshirani96}.
After generic quadratic program solvers, new dedicated solvers were proposed based on iterative reweighted least-squares (IRLS) \citep{Grandvalet98}, followed by LARS \citep{Efron_Hastie_Johnstone_Tibshirani04}, a homotopy method computing the full Lasso path\footnote{The Lasso path is the set of solutions of \autoref{pb:lasso} as $\lambda$ varies in $(0,\infty)$.}.
The LARS solver helped popularize the Lasso, yet the algorithm suffers from stability issues and can be very slow for worst case situations \citep{Mairal_Yu12}.
General purpose solvers became popular for Lasso-type problems with the introduction of the iterative soft thresholding algorithm (ISTA, \citealt{Daubechies2004}), an instance of forward-backward splitting \citep{Combettes2005}.
These algorithms became standard in signal and image processing, especially when accelerated (FISTA, \citealt{Beck_Teboulle09}).

In parallel, proximal coordinate descent has proven particularly relevant for the Lasso in statistics.
Early theoretical results were proved by \citet{Tseng93,Sardy_Bruce_Tseng00}, before it became the standard solver of the widely distributed packages \glmnet{} in \texttt{R} and \texttt{scikit-learn} in Python.
For further improvements, some solvers exploit the sparsity of $\theta^*$, trying to identify its support to reduce the problem size.
Best performing variants of this scheme are screening rules  (\eg{} \cite{Ghaoui_Viallon_Rabbani2010,Bonnefoy_Emiya_Ralaivola_Gribonval15,Ndiaye_Fercoq_Gramfort_Salmon17}) and working/active sets %
(\eg{} \citealt{johnson2015blitz,Massias_Gramfort_Salmon2018}), including strong rules \citep{Tibshirani2012}.

While reviews of Lasso solvers have already been performed \citep[Sec. 8.1]{Bach_Jenatton_Mairal_Obozinski12}, they are limited to certain implementation and design choices, but also naturally lack comparisons with more recent solvers and modern hardware, hence drawing biased conclusions.

\begin{figure}[!htbp]
    \centering
    \hspace{6mm}\includegraphics[width=0.8\figwidth]{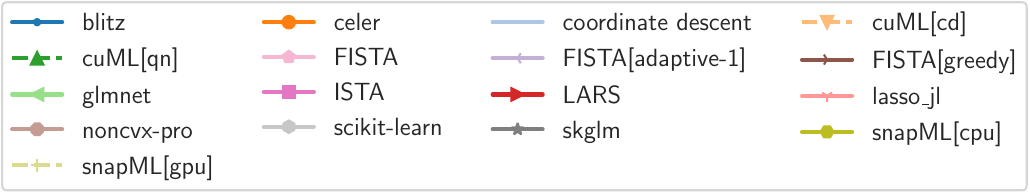}
    \includegraphics[width=0.99\figwidth]{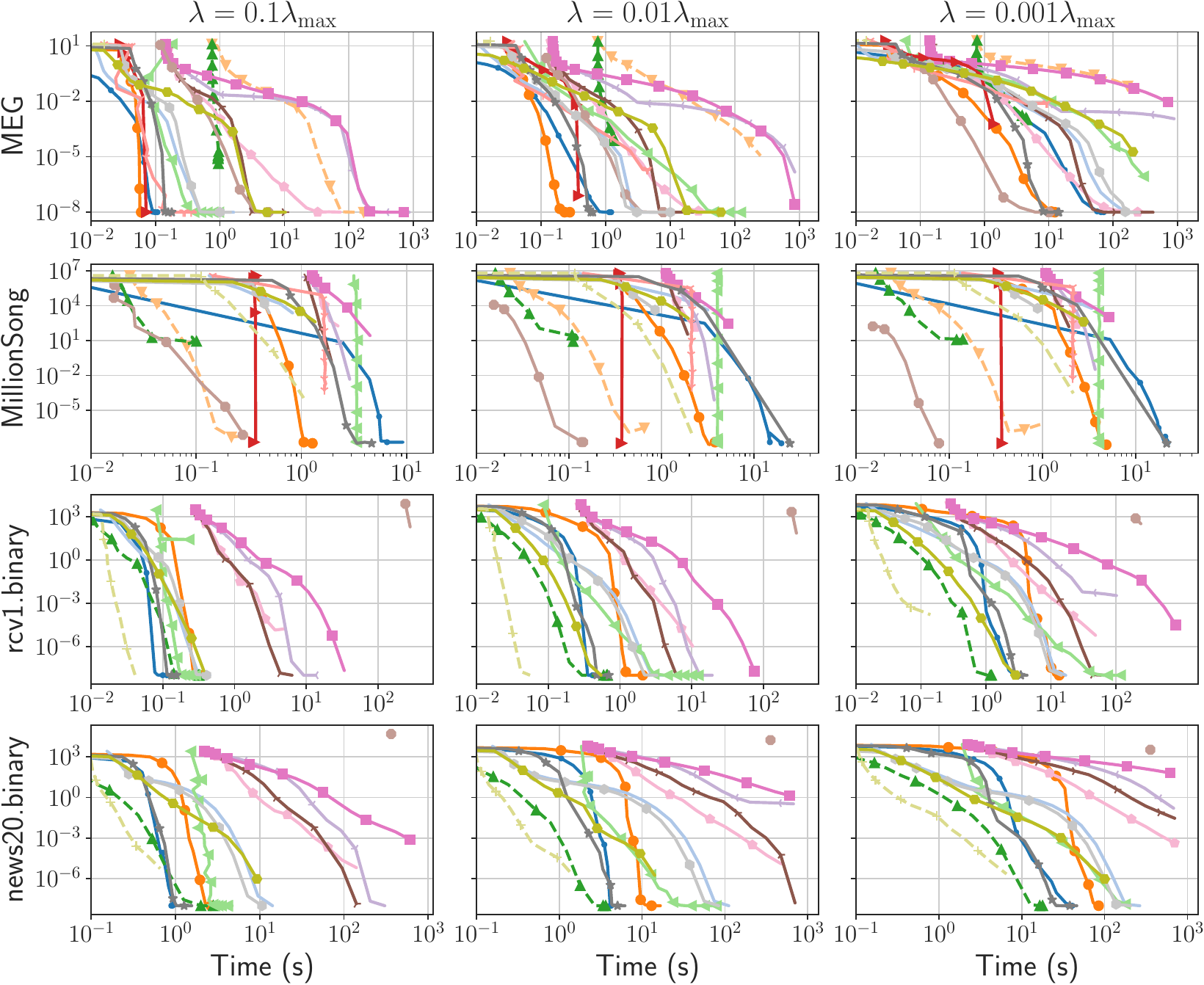}
    \caption{
        Benchmark for the Lasso, on 17 solvers, 4 datasets (\emph{rows}), and 3 variants of the \texttt{Objective} (\emph{columns}) with decreasing regularization $\lambda$.
        The curves display the suboptimality of the objective function, $f(\theta^t) - f(\theta^*)$, as a function of time.
    }
    \label{fig:lasso_leukemia_meg_rcv1_news20}\vskip-1em
\end{figure}

\looseness=-1
\repo{The code for the benchmark is available at \url{https://github.com/benchopt/benchmark_lasso/}.}
Results obtained on 4 datasets, with 9 standard packages and some custom reimplementations, possibly leveraging GPU hardware, and 17 different solvers written in \Python{}/\Numba{}/\Cython{}, \texttt{R}, \Julia{} or \texttt{C++} (\autoref{table:algo-lasso-benchmark}) are presented in \autoref{fig:lasso_leukemia_meg_rcv1_news20}.
All solvers use efficient numerical implementations, possibly leveraging calls to \BLAS, precompiled code in \Cython{} or just-in-time  compilation with \Numba{}.

The different parameters influencing the setup are
\begin{itemize}[topsep=0pt,itemsep=1ex,partopsep=0ex,parsep=0ex,leftmargin=3ex]
    \item the regularization strength $\lambda$, controlling the sparsity of the solution, parameterized as a fraction of $\lambda_{\max} = \norm{X^\top y}_\infty$ (the minimal hyperparameter such that $\theta^* = 0$),
    \item the dataset dimensions: \emph{MEG} has small $n$ and medium $p$; \emph{rcv1.binary} has medium $n$ and $p$; \emph{news20.binary} has medium $n$ and very large $p$ while \emph{MillionSong} has very large $n$ and small $p$ (\autoref{table:summary_data_lasso}).
\end{itemize}

\myparagraph{Results}
\autoref{fig:lasso_leukemia_meg_rcv1_news20} presents the result of the benchmark on the Lasso, in terms of objective suboptimality $f(\theta^t) - f(\theta^*)$ as a function of time.

Similarly to \autoref{sec:logreg}, the GPU solvers obtain good performances in most settings, but their advantage is less clear.
A consistent finding across all settings is that coordinate descent-based methods outperform full gradient ones (ISTA and FISTA, even restarted), and are improved by the use of working set strategies (\blitz{}, \celer{}, \skglm{}, \glmnet{}).
This observation is even more pronounced when the regularization parameter is large, as the solution is sparser.

When observing the influence of the dataset dimensions, we observe 3 regimes.
When $n$ is small (\emph{MEG}), the support of the solution is small and coordinate descent, LARS and \noncvxpro{} perform the best.
When $n$ is much larger than $p$ (\emph{MillionSong}), \noncvxpro{} clearly outperforms other solvers, and working set methods prove useless.
Finally, when $n$ and $p$ are large (\emph{rcv1.binary}, \emph{news20.binary}), CD and working sets vastly outperforms the rest while \noncvxpro{} fails, as it requires solving a linear system of size $\min(n, p)$.
We note that this setting was not tested in the original experiment of \citet{Poon_21}, which highlights the need for extensive, standard experimental setups.

When the support of the solution is small (either small $\lambda$, either small $n$ since the Lasso solution has at most $n$ nonzero coefficients), LARS is a competitive algorithm.
We expect this to degrade when $n$ increases, but as the LARS solver in \texttt{scikit-learn} does not support sparse design matrices we could not include it for \emph{news20.binary} and \emph{rcv1.binary}.

This benchmark is the first to evaluate solvers across languages, showing the competitive behavior of \lassojl{} and \glmnet{} compared to \Python{} solvers.
Both solvers have a large initialization time, and then converge very fast.
To ensure that the benchmark is fair, even though the \Benchopt{} library is implemented in \Python{}, we made sure to ignore conversion overhead, as well as just-in-time compilation cost.
We also checked the timing's consistency with native calls to the libraries.

Since the Lasso is massively used for it feature selection properties, the speed at which the solvers identify the support of the solution is also an important performance measure.
Monitoring this with \Benchopt{} is straightforward, and a figure reporting this benchmark is in \autoref{app:sec:lasso}. %
\newcommand{\weights}{\theta}
\newcommand{\cmark}{\ding{51}}%
\newcommand{\xmark}{\ding{55}}%

\section{Third example: How standard is a benchmark on ResNet18?}
\label{sec:resnet18}

As early successes of deep learning have been focused on computer vision tasks~\citep{Krizhevsky2012}, image classification has become a \emph{de facto} standard to validate novel methods in the field.
Among the different network architectures, ResNets~\citep{He2016} are extensively used in the community as they provide strong and versatile baselines~\citep{xie2017aggregated, tan2019efficientnet, dosovitskiy2021an, brock2021high, liu2022convnet}.
While many papers present results with such model on classical datasets, with sometimes extensive ablation studies~\citep{He2019,Wightman2021,Bello2021,Schmidt2021}, the lack of standardized codebase and missing implementation details makes it hard to replicate their results.

\repo{The code for the benchmark is available at \url{https://github.com/benchopt/benchmark_resnet_classif/}.}
We provide a cross-dataset --\emph{SVHN}, \citet{Netzer2011}; \emph{MNIST}, \citet{lecun2010mnist} and \emph{CIFAR-10}, \citet{Krizhevsky09learningmultiple}-- and cross-framework --\TensorFlow{}/\Keras{}, \citet{tensorflow2015-whitepaper,chollet2015keras}; \PyTorch{}, \citet{NEURIPS2019_9015}-- evaluation of the training strategies for image classification with ResNet18 (see \autoref{app:resnet} for details on architecture and datasets).
We train the network by minimizing the cross entropy loss relatively to the weights $\theta$ of the model.
Contrary to logistic regression and the Lasso, this problem is non-convex due to the non-linearity of the model $f_\theta$.
Another notable difference is that we report the evolution of the test error rather than the training loss.

\rebuttal{Because we chose to monitor the test loss,} the \texttt{Solvers} are defined as the combination of an optimization algorithm, its hyperparameters, the learning rate~(LR) and weight decay schedules, and the data augmentation strategy.
\rebuttal{
This is in contrast to a case where we would monitor the train loss, and therefore make the LR and weight decay schedules, as well as the data augmentation policy, part of the objective.
}
We focus on 2 standard methods: stochastic gradient descent~(SGD) with momentum and Adam~\citep{Kingma2015}, as well as a more recently published one: Lookahead~\citep{zhang2019lookahead}.
The LR schedules are chosen among fixed LR, step LR\footnote{decreasing the learning rate by a factor 10 at mid-training, and again at 3/4 of the training}, and cosine annealing~\citep{DBLP:conf/iclr/LoshchilovH17}.
We also consider decoupled weight decay for Adam~\citep{loshchilov2018decoupled}, and coupled weight decay (\ie $\ell_2$-regularization) for SGD.
Regarding data augmentation,  we use random cropping for all datasets and add horizontal flipping only for \emph{CIFAR-10}, as the digits datasets do not exhibit a mirror symmetry.
We detail the remaining hyperparameters in \autoref{tab:hp-resnet}\rebuttal{, and discuss their selection as well as their sensitivity in \autoref{app:resnet}}. %

\myparagraph{Aligning cross-framework implementations}
Due to some design choices, components with the same name in the different frameworks do not have the same behavior.
For instance, when it comes to applying weight decay, \PyTorch{}'s SGD uses coupled weight decay, while in \TensorFlow{}/\Keras{} weight decay always refers to decoupled weight decay.
These two methods lead to significantly different performance and it is not straightforward to apply coupled weight decay in a post-hoc manner in \TensorFlow{}/\Keras{} (see further details in \autoref{sec:diff-tf-pt}).
We conducted an extensive effort to align the networks implementation in different frameworks using unit testing to make the conclusions of our benchmarks independent of the chosen framework. %
We found additional significant differences (reported in \autoref{tab:diff-tf-pt}) in the initialization, the batch normalization, the convolutional layers and the weight decay scaling. %

\myparagraph{Results}
\begin{figure}[t]
    \centering
    \includegraphics[width=1\figwidth]{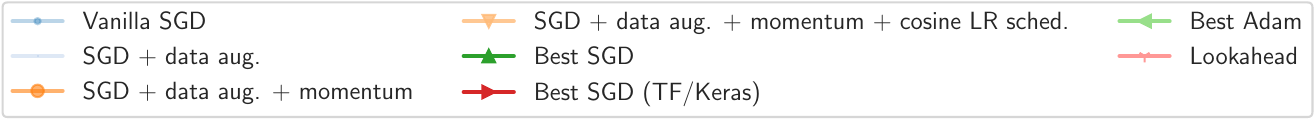}
    \includegraphics[width=0.99\figwidth]{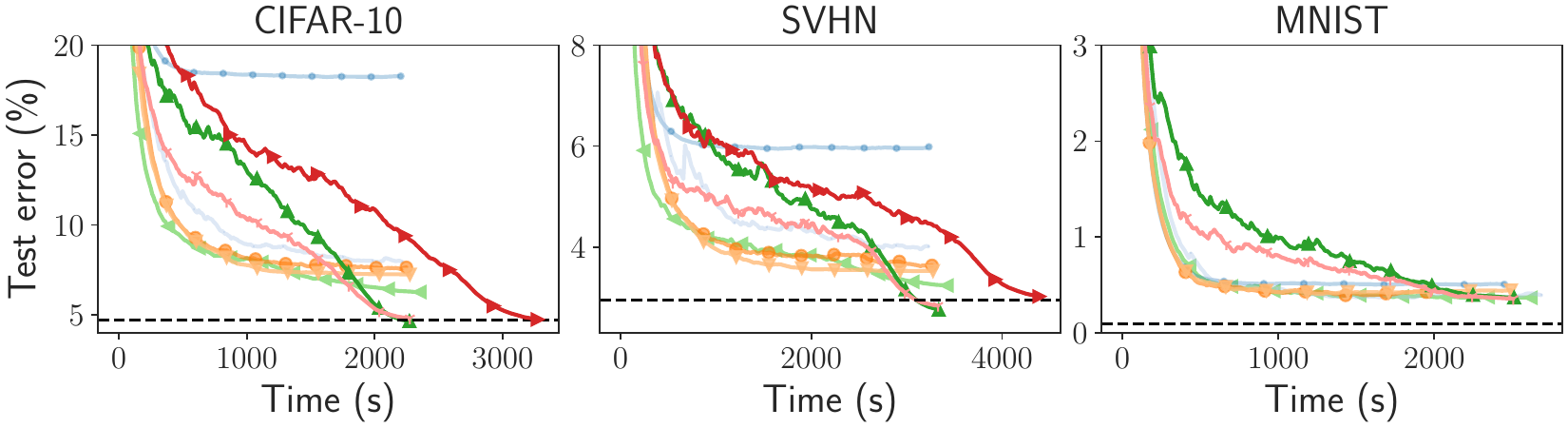}
    \caption{\textbf{ResNet18 image classification benchmark with \PyTorch{} \texttt{Solvers}.} The best SGD configuration features data augmentation, momentum, cosine learning rate schedule and weight decay. In dashed black is the state of the art for the corresponding datasets with a ResNet18 measured by \citet{zhang2019lookahead} for \emph{CIFAR-10}, by \citet{zheng2021regularizing} for \emph{SVHN} with a PreAct ResNet18,
    by \href{https://paperswithcode.com/sota/image-classification-on-mnist}{PapersWithCode} for \emph{MNIST} with all networks considered. Off-the-shelf ResNet implementations in \TensorFlow/\Keras{} do not support images smaller than $32\times32$ and is hence not shown for \emph{MNIST}.\repo{Curves are exponentially smoothed.}}
    \label{fig:resnet18_sgd_torch}\vskip-1.1em
\end{figure}
The results of the benchmark are reported in \autoref{fig:resnet18_sgd_torch}.
Each graph reports the test error relative to time, with an ablation study on the solvers parameters.
Note that we only report selected settings for clarity but that we run every possible combinations.\footnote{The results are available online as a \href{https://benchopt.github.io/results/benchmark_resnet_classif_benchmark_resnet_classif_benchopt_run_2022-05-03_10h31m54.html}{user-friendly interactive \HTML{} file}}.

Firstly, reaching the state of the art for a vanilla ResNet18 is not straightforward.
On the popular website \href{https://paperswithcode.com/}{Papers with code} it has been so far underestimated.
It can achieve 4.45\% and 2.65\% test error rates on \emph{CIFAR-10} and \emph{SVHN} respectively (compared to 4.73\% and 2.95\% -- for a PreAct ResNet18 -- before that).
Our ablation study shows that a variety of techniques is required to reach it.
The most significant one is an appropriate data augmentation strategy, which lowers the error rate on \emph{CIFAR-10} from about 18\% to about 8\%.
The second most important one is weight decay, but it has to be used in combination with a proper LR schedule, as well as momentum.
While these techniques are not novel, they are regularly overlooked in baselines, resulting in underestimation of their performance level.

This reproducible benchmark not only allows a researcher to get a clear understanding of how to achieve the best performances for this model and datasets, but also provides a way to reproduce and extend these performances.
In particular, we also include in this benchmark the original implementation of Lookahead~\citep{zhang2019lookahead}.
We confirm that it slightly accelerates the convergence of the Best SGD, even with a cosine LR schedule -- a setting that had not been studied in the original paper.

Our benchmark also evaluates the relative computational performances of the different frameworks.
We observe that \texttt{PyTorch-Lightning} is significantly slower than the other frameworks we tested, in large part due to their callbacks API.
We also notice that our \TensorFlow/\Keras{} implementation is significantly slower ($\approx$ 28\%) than the \PyTorch{} ones, despite following the best practices and our profiling efforts.
Note that we do not imply that \TensorFlow{} is intrinsically slower than \PyTorch{}, but a community effort is needed to ensure that the benchmark performances are framework-agnostic.

\begin{figure}[h!]
    \centering
    \includegraphics[width=1.0\figwidth]{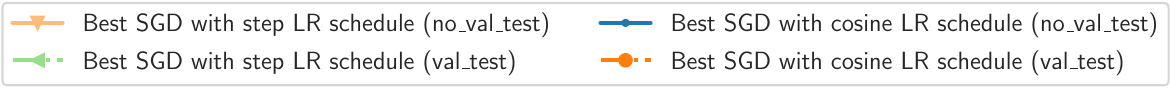}
    \includegraphics[width=0.99\figwidth]{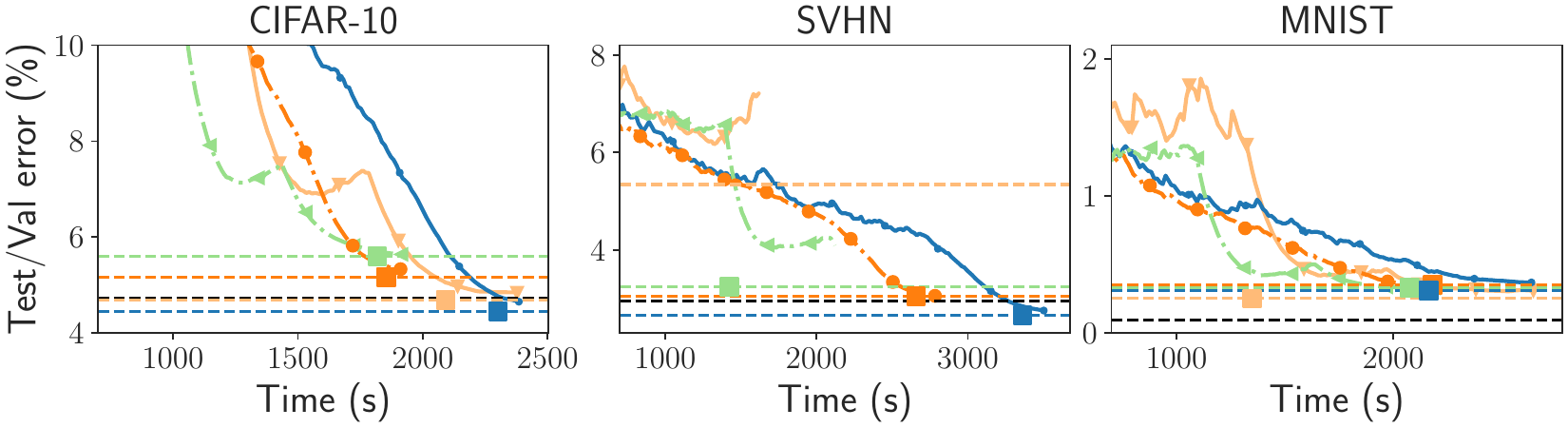}
    \caption{\textbf{ResNet18 image classification benchmark with a validation split.} In dashed black is the state of the art (see caption of \autoref{fig:resnet18_sgd_torch} for more details). In addition, we show in colored horizontal dashed lines, the test results for early stopping on the validation and on the test set for the different solvers, the square mark indicating the moment this stopping would happen. The curves for the train-val splits show the exponentially smoothed median results for five different random seeds.}
    \label{fig:resnet18_sgd_torch_val}\vskip-1em
\end{figure}
A recurrent criticism of such benchmarks is that only the best test error is reported.
In \autoref{fig:resnet18_sgd_torch_val}, we measure the effect of using a train-validation-test split, by keeping a fraction of the training set as a validation set.
The splits we use are detailed in \autoref{tab:resnet-datasets}.
Our finding is that the results of the ablation study do not change significantly when using such procedure, even though their validity is reinforced by the use of multiple trainings.
Yet, a possible limitation of our findings is that some of the hyperparameters we used for our study, coming from the \href{https://github.com/kuangliu/pytorch-cifar}{PyTorch-CIFAR GitHub repository}, may have been tuned while looking at the test set.
\section{Conclusion and future work}
We have introduced \Benchopt{}, a library that makes it easy to collaboratively develop fair and extensive benchmarks of optimization algorithms, which can then be seamlessly published, reproduced, and extended.
In the future, we plan on supporting the creation of new benchmarks, that could become the standards the community builds on.
This work is part of a wider effort to improve reproducibility of machine learning results.
It aims to contribute to raising the standard of numerical validation for optimization, which is pervasive in the statistics and ML community as well as for the experimental sciences that rely more and more on these tools for research.

\section{Acknowledgements}\label{app:sec:acknowledgments}

It can not be stressed enough how much the \Benchopt{} library relies on contributions from the community and in particular the Python open source ecosystem.
In particular, it could not exist without the libraries mentioned in \autoref{sec:app:software}.

This work was granted access to the HPC resources of IDRIS under the allocation 2022-AD011011172R2 and 2022-AD011013570 made by GENCI, which was used to run all the benchmarks.
MM also gratefully acknowledges the support of the Centre Blaise Pascal's IT test platform at ENS de Lyon (Lyon, France) for Machine Learning facilities.
The platform operates the SIDUS solution \citep{quemener2013sidus}.

TL, CFD and JS contributions were supported by the Chaire IA CaMeLOt (ANR-20-CHIA-0001-01).
AG, EL and TM contributions were supported by the Chaire IA ANR BrAIN (ANR-20-CHIA-0016).
BMa contributions were supported by a grant from Digiteo France.
MD contributions were supported by a public grant overseen by the French National Research Agency (ANR) through the program UDOPIA, project funded by the ANR-20-THIA-0013-01 and DATAIA convergence institute  (ANR-17-CONV-0003).
BN work was supported by the Télécom Paris's Chaire DSAIDIS (Data Science \& Artificial Intelligence for Digitalized Industry Services).
BMo contributions were supported by a grant from the Labex MILYON.
{
\setlength\bibitemsep{3\itemsep}  %
\small
\printbibliography
}

\newpage
\section*{Checklist}

\begin{enumerate}

\item For all authors...
\begin{enumerate}
  \item Do the main claims made in the abstract and introduction accurately reflect the paper's contributions and scope?
    \answerYes{}
  \item Did you describe the limitations of your work?
    \answerYes{}
  \item Did you discuss any potential negative societal impacts of your work?
    \answerYes{}
  \item Have you read the ethics review guidelines and ensured that your paper conforms to them?
    \answerYes{}
\end{enumerate}

\item If you are including theoretical results...
\begin{enumerate}
  \item Did you state the full set of assumptions of all theoretical results?
    \answerNA{}
	\item Did you include complete proofs of all theoretical results?
    \answerNA{}
\end{enumerate}

\item If you ran experiments...
\begin{enumerate}
  \item Did you include the code, data, and instructions needed to reproduce the main experimental results (either in the supplemental material or as a URL)?
    \answerYes{}
  \item Did you specify all the training details (e.g., data splits, hyperparameters, how they were chosen)?
    \answerYes{} These are specified in Appendix, on a per-benchmark basis.
	\item Did you report error bars (e.g., with respect to the random seed after running experiments multiple times)?
    \answerNo{} Error bars are not reported for clarity, but \Benchopt{} allows this in particular in html versions of the plots that can be found in \url{https://benchopt.github.io/results/preprint_results.html}.
	\item Did you include the total amount of compute and the type of resources used (e.g., type of GPUs, internal cluster, or cloud provider)?
    \answerYes{}
\end{enumerate}

\item If you are using existing assets (e.g., code, data, models) or curating/releasing new assets...
\begin{enumerate}
  \item If your work uses existing assets, did you cite the creators?
    \answerYes{}
  \item Did you mention the license of the assets?
    \answerYes{} In the introduction.
  \item Did you include any new assets either in the supplemental material or as a URL?
    \answerYes{}
  \item Did you discuss whether and how consent was obtained from people whose data you're using/curating?
    \answerNA{}
  \item Did you discuss whether the data you are using/curating contains personally identifiable information or offensive content?
    \answerNA{}
\end{enumerate}

\item If you used crowdsourcing or conducted research with human subjects...
\begin{enumerate}
  \item Did you include the full text of instructions given to participants and screenshots, if applicable?
    \answerNA{}
  \item Did you describe any potential participant risks, with links to Institutional Review Board (IRB) approvals, if applicable?
    \answerNA{}
  \item Did you include the estimated hourly wage paid to participants and the total amount spent on participant compensation?
    \answerNA{}
\end{enumerate}

\end{enumerate}

\newpage
\appendix
\def\thetable{\thesection.\arabic{table}}
\counterwithin{table}{section}
\counterwithin{figure}{section}

\clearpage{}%
\section{Software ecosystem acknowledgement}
\label{sec:app:software}

The command line interface and API use the \texttt{click}, \texttt{pyyaml} and \texttt{psutil} \citep{psutil} libraries.

Numerical computations involve \texttt{numpy} \citep{harris2020array} and \texttt{scipy} \citep{2020SciPy-NMeth}.
For cross-language processing, we used \texttt{rpy2} for calling \texttt{R}~\citep{Rlang} libraries and \texttt{PyJulia} for interfacing with \texttt{Julia}~\citep{Julia}.
The benchmark runs extensively use \texttt{joblib}, \texttt{loky} \citep{loky} and \texttt{submitit} for parallelization.

The results are stored and processed for visualizations using \texttt{pandas}~\citep{pandas}, \texttt{matplotlib} \citep{matplotlib} for static rendering, \texttt{mako} and \texttt{plotly} \citep{plotly} for interactive webpages. The participative results website relies partially on \texttt{pygithub}.

Our documentation is generated by multiple \texttt{sphinx}-based \citep{sphinx} libraries (\texttt{sphinx-bootstrap-theme}, \texttt{sphinx-click}, \texttt{sphinx-gallery} \citep{sphinxgallery} and \texttt{sphinx-prompt}), and  also the \texttt{numpydoc} and \texttt{pillow} \citep{clark2015pillow} libraries.
\clearpage{}%
\clearpage{}%
\section{A complete Benchmark example: \texttt{Objective}, \texttt{Dataset} and \texttt{Solver} classes for Ridge regression}\label{app:sec:code}

Here, we provide code examples for a simple benchmark on  Ridge regression.
The Ridge regression -- also called $\ell_2$-regularized least-squares or Tikhonov regression -- is a popular method to solve least-square problems in the presence of noisy observations or correlated features.
The problem reads:
\begin{problem}
\label{pb:ridge}
    \min_{\theta} \frac12\|y - X\theta\|_2^2 + \frac{\lambda}{2}\|\theta\|_2^2 \enspace ,
\end{problem}
where $X \in \mathbb{R}^{n\times p}$ is a design matrix, $y\in \mathbb{R}^n$ is the target vector and $\lambda$ is the regularization parameter.
This problem is strongly convex and many methods can be used to solve it.
Direct computation of the close form solution $\theta^* = (X^\top X + \lambda Id)^{-1}X^\top y$ can be obtained using matrix factorization methods such as Cholesky decomposition or the SVD~\citep{press2007numerical} or iterative linear solver such as Conjugate-Gradient~\citep{Liu1989}.
One can also resort on first order methods such as gradient descent, coordinate descent (known as the  Gauss-Seidel method in this context), or their stochastic variant.

\repo{The code for the benchmark is available at \url{https://github.com/benchopt/benchmark_ridge/}.}
The following code snippets are provided in the documentation as a template for new benchmarks.

\subsection{\texttt{Objective} class}\label{app:sec:objective_code}

The \texttt{Objective} class is the central part of the benchmark, defining the objective function.
This class allows us to monitor the quantities of interest along the iterations of the solvers, amongst which the objective function value.
An \texttt{Objective} class should define 3 methods:

\begin{itemize}[topsep=0pt,itemsep=1ex,partopsep=0ex,parsep=0ex,leftmargin=3ex]
    \item \texttt{set\_data(**data)}: it allows specifying the nature of the data used in the benchmark.
    The data is passed as a dictionary of Python variables, so no constraint is enforced to what can be passed here.
    \item \texttt{compute(theta)}: it allows evaluating the objective function for a given value of the iterate, here called $\theta$.
    This method should take only one parameter, the output returned by the \texttt{Solver}.
    All other parameters should be stored in the class with the \texttt{set\_data} method.
    The compute function should return a float (understood as the objective value) or a dictionary.
    If a dictionary is returned it should contain a key called \texttt{value} (the objective value) and all other keys should correspond to float values allowing tracking more than one quantity of interest (e.g. train and test errors).
    \item \texttt{to\_dict()}: a method that returns a dictionary to be passed to the \texttt{set\_objective()} method of a \texttt{Solver}.
\end{itemize}
An \texttt{Objective} class needs to inherit from a base class, \texttt{benchopt.BaseObjective}.
Below is the implementation of the Ridge regression \texttt{Objective} class.

\begin{python}
from benchopt import BaseObjective

class Objective(BaseObjective):
    name = "Ridge regression"
    parameters = {"reg": [0.1, 1, 10]}

    def __init__(self, reg=1):
        self.reg = reg

    def set_data(self, X, y):
        self.X, self.y = X, y

    def compute(self, theta):
        res = self.y - self.X @ theta
        return .5 * res @ res + 0.5 * self.reg * theta @ theta

    def to_dict(self):
        return dict(X=self.X, y=self.y, reg=self.reg)

\end{python}

\subsection{\texttt{Dataset} class}

A \texttt{Dataset} class defines data to be passed to the \texttt{Objective}.
More specifically, a \texttt{Dataset} class should implement one method:
\begin{itemize}[topsep=0pt,itemsep=1ex,partopsep=0ex,parsep=0ex,leftmargin=3ex]
    \item \texttt{get\_data()}: A method outputting a dictionary that can be passed as keyword arguments \texttt{**data} to the \texttt{set\_data} method of the \texttt{Objective}.
\end{itemize}
A \texttt{Dataset} class also needs to inherit from a base class, \texttt{benchopt.BaseDataset}.

If a \texttt{Dataset} requires some packages to function, \Benchopt{} allows listing some requirements.
The necessary packages should be available via \texttt{conda} or \texttt{pip}.

Below is an example of a \texttt{Dataset} definition using the \texttt{libsvmdata} library, which exposes datasets from \texttt{libsvm}, such as \emph{leukemia}, \emph{bodyfat} and \emph{gisette} -- described in \autoref{table:summary_data_ridge}.

\begin{python}
from benchopt import BaseDataset
from benchopt import safe_import_context

# This context allow to manipulate the Dataset object even if
# libsvmdata is not installed. It is used in `benchopt install`.
with safe_import_context() as import_ctx:
    from libsvmdata import fetch_libsvm

class Dataset(BaseDataset):
    name = "libsvm"
    install_cmd = "conda"
    requirements = ["libsvmdata"]
    parameters = {"dataset": ["bodyfat", "leukemia", "gisette"]}

    def __init__(self, dataset="bodyfat"):
        self.dataset = dataset

    def get_data(self):
        X, y = fetch_libsvm(self.dataset)
        return dict(X=self.X, y=self.y)
\end{python}

\subsection{\texttt{Solver} class}

A \texttt{Solver} class must define three methods:
\begin{itemize}[topsep=0pt,itemsep=1ex,partopsep=0ex,parsep=0ex,leftmargin=3ex]
    \item \texttt{set\_objective(**objective\_dict)}: This method will be called with the dictionary \texttt{objective\_dict} returned by the method \texttt{to\_dict} from the \texttt{Objective}. The goal of this method is to provide all necessary information to the \texttt{Solver} so it can optimize the objective function.
    \item \texttt{run(stop\_value)}: This method takes only one parameter that controls the stopping condition of the \texttt{Solver}. Typically this is either a number of iterations \texttt{n\_iter} or a tolerance parameter \texttt{tol}. Alternatively, a callback function that will be called at each iteration can be passed. The callback should return \texttt{False} once the computation should stop. The parameter \texttt{stop\_value} is controlled by the \texttt{stopping\_strategy}, see below for details.
    \item \texttt{get\_result()}: This method returns a variable that can be passed to the compute method from the \texttt{Objective}. This is the output of the \texttt{Solver}.
\end{itemize}

If a Python \texttt{Solver} requires some packages such as \texttt{scikit-learn}, \Benchopt{} allows listing some requirements.
The necessary packages must be available via \texttt{conda} or \texttt{pip}.

Below is a simple \texttt{Solver} example using \texttt{scikit-learn} implementation of Ridge regression with different optimization algorithms.

\newpage
\begin{python}
from benchopt import BaseSolver
from benchopt import safe_import_context

# This context allow to manipulate the Solver object even if
# scikit-learn is not installed. It is used in `benchopt install`.
with safe_import_context() as import_ctx:
    from sklearn.linear_model import Ridge

class Solver(BaseSolver):
    name = "scikit-learn"
    install_cmd = "conda"
    requirements = ["scikit-learn"]
    parameters = {
        "alg": ["svd", "cholesky", "lsqr", "sparse_cg", "saga"],
    }

    def __init__(self, alg="svd"):
        self.alg = alg

    def set_objective(self, X, y, reg=1):
        self.X, self.y  = X, y
        self.clf = Ridge(
            fit_intercept=False, alpha=reg, solver=self.alg,
            tol=1e-10
        )

    def run(self, n_iter):
        self.clf.max_iter = n_iter + 1
        self.clf.fit(self.X, self.y)

    def get_result(self):
        return self.clf.coef_
\end{python}

\subsection{Results from the benchmark}

\myparagraph{Descriptions of datasets}
\autoref{table:summary_data_ridge} describes the datasets used in this benchmarks.

\begin{table}[h]
  \centering
  \caption{List of the datasets used in Ridge regression in \autoref{app:sec:code}}
      \begin{tabular}{llS[table-format=8.0]S[table-format=7.0]S[table-format=1.1e-1,scientific-notation=true]}
        \toprule
        \textbf{Datasets} & \textbf{References} &{\textbf{Samples ($\mathbf{n}$)}} & {\textbf{Features ($\mathbf{p}$)}} \\
        \midrule
        \emph{leukemia} &\citet{leukemia} & 38 & 7129  \\
        \emph{bodyfat} &\citet{madelon} & 252 & 8   \\
        \emph{gisette} &\citet{madelon} & 6000 & 5000  \\
        \bottomrule
    \end{tabular}
    \label{table:summary_data_ridge}
\end{table}

We also run the solvers on the simulated data described bellow.

\myparagraph{Generation process for \texttt{simulated}
dataset}
\label{app:simulated_dataset}
We generate a linear regression scenario with decaying correlation for the design matrix, \ie the ground-truth covariance matrix is a Toeplitz matrix, with each element $\Sigma_{ij} = \rho^{|i-j|}$.
As a consequence, the generated features have 0 mean, a variance of 1, and the correlation structure as:
\begin{equation}
            \mathbb E[X_i] = 0~, \quad \mathbb E[X_i^2] = 1  \quad
        \text{and} \quad \mathbb E[X_iX_j] = \rho^{|i-j|} \ .
\end{equation}
Our simulation scheme also includes the parameter $\texttt{density}=0.2$ that controls the proportion of non-zero elements in $\theta^*$.
The target vector is generated according to linear relationship with Gaussian noise:
\begin{equation*}
     y = X \theta^* + \varepsilon \ ,
\end{equation*}
such that the signal-to-noise ratio is $\texttt{snr} = \frac{\|X \theta^*\|_2}{\|\varepsilon\|_2}$.

We use a signal-to-noise ratio $\texttt{snr}=3$, a correlation $\rho$ of $0$ or $0.6$ with $n=500$ samples and $p=1000$ features.

\myparagraph{Description of the solvers} \autoref{table:algo-ridge-benchmark} describes the different solvers compared in this benchmark.

\begin{table}[h]
  \centering
  \footnotesize
  \caption{List of solvers used in the Ridge benchmark in \autoref{app:sec:code}}
  \addtolength{\tabcolsep}{-1pt}
\begin{tabular}{l l p{3cm} l}
  \toprule
  \textbf{Solver} & \textbf{References} & \textbf{Description} & \textbf{Language}\\
  \midrule
  \texttt{GD} &\citet{Boyd2004}
                 & Gradient Descent
                 & \texttt{Python}\\
  \texttt{Accelerated GD} &\citet{nesterov1983accelerated}
                 & Gradient Descent + acceleration
                 & \texttt{Python}\\
 \texttt{scikit-learn[svd]} & \citet{Pedregosa_11} & SVD (Singular Value Decomposition) & \texttt{Python} (\texttt{Cython})\\
  \texttt{scikit-learn[cholesky]} & \citet{Pedregosa_11} & Cholesky decomposition & \texttt{Python} (\texttt{Cython})\\
  \texttt{scikit-learn[lsqr]} & \citet{Pedregosa_11} & regularized least-squares & \texttt{Python} (\texttt{Cython})\\
  \texttt{scikit-learn[saga]} & \citet{Pedregosa_11} & SAGA (Varianced reduced stochastic method) & \texttt{Python} (\texttt{Cython})\\
    \texttt{scikit-learn[cg]} & \citet{Pedregosa_11} & Conjugate Gradient & \texttt{Python} (\texttt{Cython})\\
  \texttt{CD} & \citet{Bertsekas99} & Cyclic Coordinate Descent & \texttt{Python} (\texttt{Numba})\\
  \texttt{lightning[cd]} & \citet{Blondel2016} & Cyclic Coordinate Descent & \texttt{Python} (\texttt{Cython})\\
  \texttt{snapML[cpu]} &\citet{Dunner_18}
                     & CD
                     & \texttt{Python}, \texttt{C++}\\
    \texttt{snapML[gpu]} &\citet{Dunner_18}
                     & CD + GPU
                     & \texttt{Python}, \texttt{C++}\\
  \bottomrule
\end{tabular}
\label{table:algo-ridge-benchmark}
\end{table}

\begin{figure}[t]
    \centering
    \includegraphics[width=.8\figwidth]{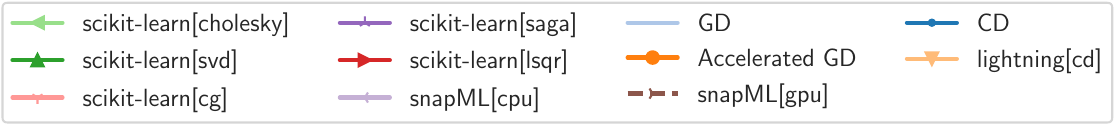}
    \includegraphics[width=0.99\figwidth]{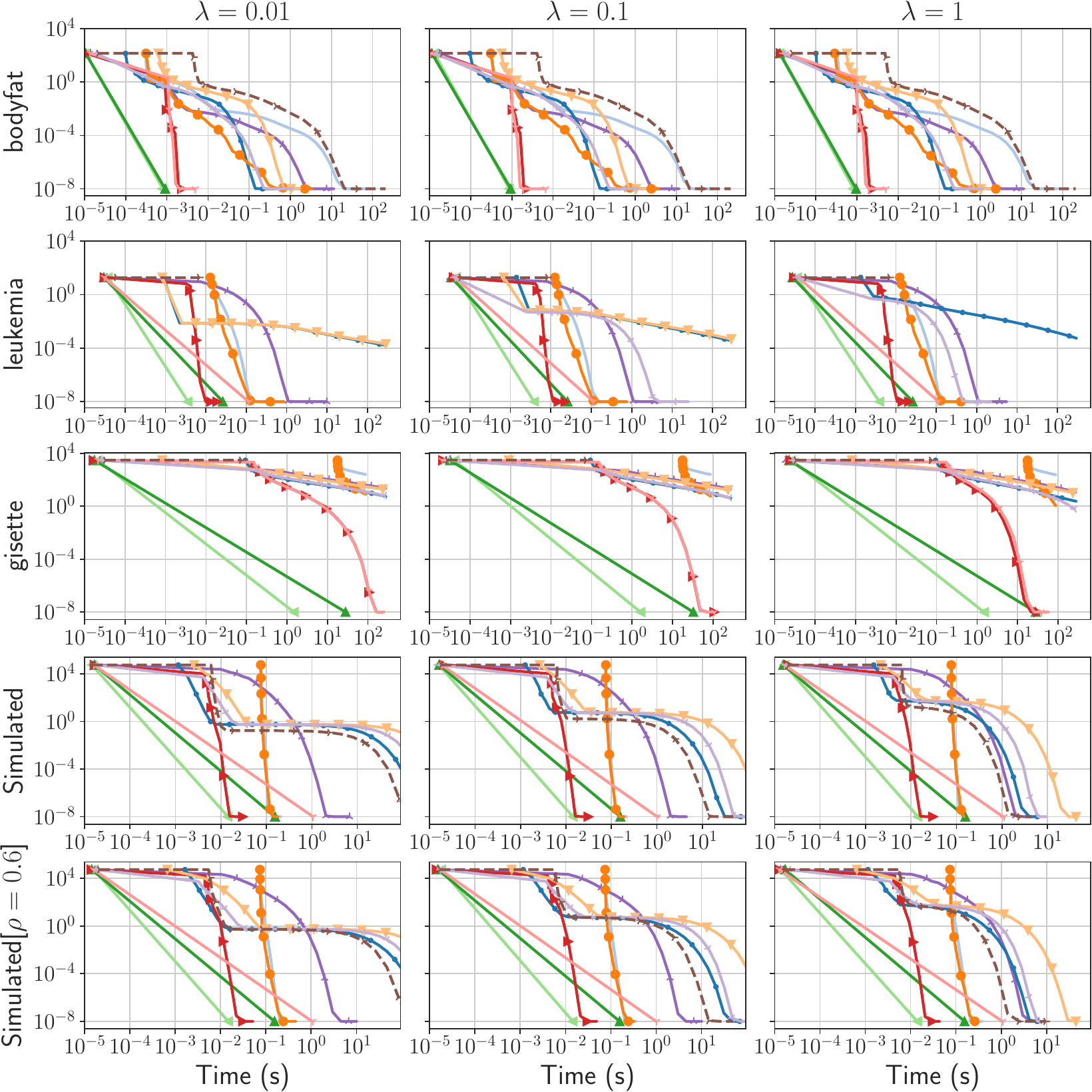}
    \caption{
        Benchmark for the Ridge regression, on 10 solvers, 5 datasets (\emph{rows}), and
        3 variants of the \texttt{Objective}
        (\emph{columns}) each with a different regularization value $\lambda \in \{ 0.01, 0.1, 1\}$.
        The curves display the suboptimality of the iterates, $f(\theta^t) - f(\theta^*)$, as a function of time.
    }
    \label{fig:ridge_medium}
\end{figure}

\myparagraph{Results} \autoref{fig:ridge_medium} presents the performance of the different methods for different values of the regularization parameter in the benchmark.
The algorithms based on the direct computation of the closed-form solution outperform iterative ones in a majority of presented datasets.
Among closed-form algorithms, the Cholesky solver converges faster.

\clearpage{}%
\clearpage{}%
\section{Design choices}
\label{app:design_choices}
\rebuttal{\Benchopt{} has made some design choices, while trying as much as possible to leave users free of customizing the behavior on each benchmark.
We detail the most important ones in this section.}

\subsection{Estimating \texorpdfstring{$\theta^*$}~  for convex problems}
\label{app:sec:best_iterate}

When the problem is convex, many solvers are guaranteed to converge to a global minimizer $\theta^*$ of the objective function $f$.
To estimate $\theta^*$ and $f(\theta^*)$, \Benchopt{} approximates  $\theta^*$ by the iterate achieving the lowest \texttt{objective\_value} among all solvers for a given \texttt{Dataset} and \texttt{Objective}.
This means that the sub-optimality plot proposed by \Benchopt{} are only valid if at least one solver has converged to the optimal solution. Else, the curves are a lower bound estimate of the sub-optimality.
In practice, for most considered convex problems, running the \texttt{Solver} for long enough ensures that $f(\theta^*)$ is correctly estimated.

\subsection{Stopping solvers}
\label{app:sec:stopping_solver}
\rebuttal{
\Benchopt{} offers many ways to stop running a solver.
The most common is to stop the solver when the objective value does not decrease significantly between iterations.
For some convex problems, we also propose to track the duality gap (which upper bounds the suboptimality), as is done for the Lasso.
For non convex problems, criteria such as gradient norm or violation of first order conditions can be used, as users do in practice.
These criteria can easily be customized.
}

\subsection{Wall-clock time versus number of iterations}
\label{app:sec:wall_clock_time}
\rebuttal{
Measuring time or iteration are two alternatives that make sense in their respective contexts. Practitioners mostly care about the time it takes to solve their problem, while researchers in mathematical optimization may want to abstract away the implementation and hardware details and only consider iteration.
The benchmarks we have presented showcase efficient implementations and are also interested in hardware and implementation differences (e.g. CPU vs GPU solvers for \autoref{sec:logreg,sec:lasso}, torch versus tensorflow for \autoref{sec:bench-vgg}), hence our focus on time.
However, \Benchopt{} does not impose a choice between the two measures: it is perfectly possible to create plots as a function of the number of iterations as evidenced for example in \autoref{app:sec:lasso_iteration}.
}\clearpage{}%
\clearpage{}%
\section{\texorpdfstring{$\ell_2$}{l2}-regularized logistic regression}
\label{app:logregl2}

\subsection{List of solvers and datasets used in the benchmark in \autoref{sec:logreg}}

\autoref{table:algo-logreg-benchmark} and \autoref{table:summary_data_logreg} respectively present the \texttt{Solvers} and \texttt{Datasets} used in this benchmark.

\begin{table}[h]
  \centering
  \footnotesize
  \caption{List of solvers used in the $\ell_2$-regularized logistic regression benchmark in \autoref{sec:logreg}}
\begin{tabular}{l p{4.1cm} p{2cm} l}
  \toprule
  \textbf{Solver} & \textbf{References} &\textbf{Description} & \textbf{Language}\\
  \midrule
  \texttt{lightning[sag]} & \citet{Blondel2016}
                          & SAG
                          &\texttt{Python} (\texttt{Cython})\\
  \texttt{lightning[saga]} &\citet{Blondel2016}
                           & SAGA
                           & \texttt{Python} (\texttt{Cython})\\
  \texttt{lightning[cd]} & \citet{Blondel2016}
                         & Cyclic Coordinate Descent
                         & \texttt{Python} (\texttt{Cython})\\
  \texttt{Tick[svrg]} & \citet{Bacry2017}
                      & Stochastic Variance Reduced Gradient
                      & \texttt{Python}, \texttt{C++}\\
  \texttt{scikit-learn[sgd]} & \citet{Pedregosa_11}
                             & Stochastic Gradient Descent
                             & \texttt{Python} (\texttt{Cython})\\
  \texttt{scikit-learn[sag]} & \citet{Pedregosa_11}
                             & SAG
                             & \texttt{Python} (\texttt{Cython})\\
  \texttt{scikit-learn[saga]} & \citet{Pedregosa_11}
                              & SAGA
                              & \texttt{Python} (\texttt{Cython})\\
  \texttt{scikit-learn[liblinear]} &\citet{Pedregosa_11}, \citet{Fan2008}
                                   & Truncated Newton Conjugate-Gradient
                                   & \texttt{Python} (\texttt{Cython})\\
  \texttt{scikit-learn[lbfgs]} & \citet{Pedregosa_11}, \citet{2020SciPy-NMeth}
                               & L-BFGS (Quasi-Newton Method)
                               & \texttt{Python} (\texttt{Cython})\\
  \texttt{scikit-learn[newton-cg]} &\citet{Pedregosa_11}, \citet{2020SciPy-NMeth}
                     & Truncated Newton Conjugate-Gradient
                     & \texttt{Python} (\texttt{Cython})\\
  \texttt{snapml[cpu]} &\citet{Dunner_18}
                             & CD
                             & \texttt{Python}, \texttt{C++}\\
  \texttt{snapml[gpu]} & \citet{Dunner_18}
                            & CD + GPU
                            & \texttt{Python}, \texttt{C++}\\
  \texttt{cuML[gpu]} & \citet{Raschka2020a}
                     & L-BFGS + GPU
                     & \texttt{Python}, \texttt{C++}\\
  \bottomrule
\end{tabular}
\label{table:algo-logreg-benchmark}
\end{table}

\begin{table}[h]
  \centering
  \caption{List of the datasets used in $\ell_2$-regularized logistic regression in \autoref{sec:logreg}}
      \begin{tabular}{llS[table-format=8.0]S[table-format=7.0]S[table-format=1.1e-1,scientific-notation=true]}
        \toprule
        \textbf{Datasets} & \textbf{References} &{\textbf{Samples ($\mathbf{n}$)}} & {\textbf{Features ($\mathbf{p}$)}} & {\textbf{Density}} \\
        \midrule
        \emph{ijcnn1} &\citet{ijcnn1} & 49990 & 22 & 0.045 \\
        \emph{madelon} &\citet{madelon} & 2000 & 500 & 0.0020  \\
        \emph{news20.binary} &\citet{news20} & 19996 & 1355191 & 0.00034 \\
        \emph{criteo} &\citet{Criteo_15}  &  45840617 & 1000000 &  0.000039 \\
        \bottomrule
    \end{tabular}
    \label{table:summary_data_logreg}
\end{table}

\subsection{Results}

\autoref{fig:logreg_l2_medium} presents the performance results for the different solvers on the different datasets using various regularization parameter values, on unscaled raw data.
We observe that when the regularization parameter $\lambda$ increases, the problem tends to become easier and faster to solve for most methods.
Also, the relative order of the method does not change significantly for the considered range of regularization.

\begin{figure}[t]
    \centering
    \includegraphics[width=.9\figwidth]{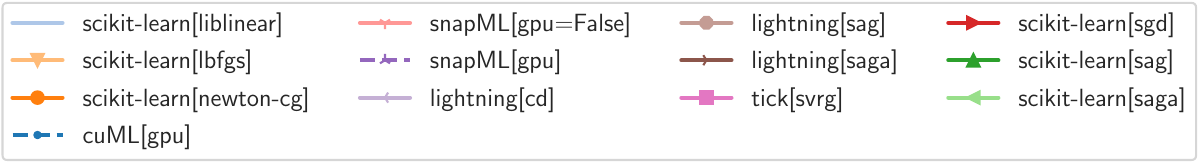}
    \includegraphics[width=0.99\figwidth]{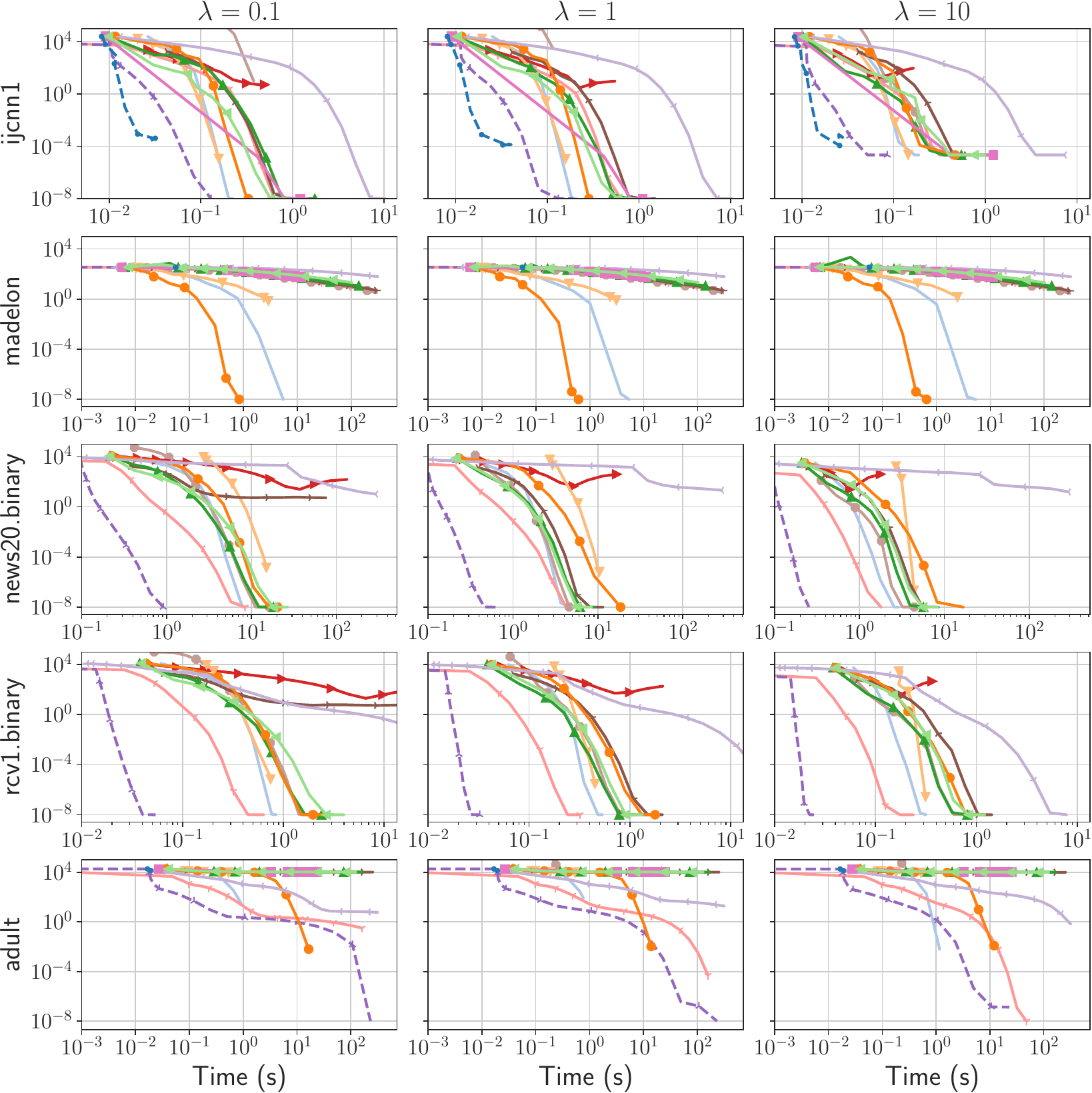}
    \caption{
        Additional benchmark for the $\ell_2$-regularized logistic regression on variants of the \texttt{Objective} (\emph{columns}) with \texttt{fit\_intercept=False}.
        The curves display the suboptimality of the iterates, $f(\theta^t) - f(\theta^*)$, as a function of time.
        The columns correspond to the objective detailed in \autoref{pb:logreg} with different value of $\lambda$: (\emph{first}) $\lambda = 0.1$, (\emph{second}) $\lambda = 1$ and (\emph{third}) $\lambda=10$.
    }
    \label{fig:logreg_l2_medium}
\end{figure}\clearpage{}%
\clearpage{}%

\section{Lasso}\label{app:sec:lasso}

\subsection{List of solvers and datasets used in the Lasso benchmark in \autoref{sec:lasso}}

\autoref{table:algo-lasso-benchmark} and \autoref{table:summary_data_lasso} respectively present the \texttt{Solvers} and \texttt{Datasets} used in this benchmark.
\begin{table}[h]
  \centering
  \footnotesize
  \caption{List of solvers used in the Lasso benchmark in \autoref{sec:lasso}}
  \addtolength{\tabcolsep}{-1pt}
\begin{tabular}{l p{4cm} p{3cm} l}
  \toprule
  \textbf{Solver} & \textbf{References} & \textbf{Description} & \textbf{Language}\\
  \midrule
  \texttt{blitz} &\citet{johnson2015blitz}
                 & CD + working set
                 & \texttt{Python}, \texttt{C++}\\
  \texttt{coordinate descent} &\citet{Friedman_10}
                              & (Cyclic) Minimization along coordinates
                              & \texttt{Python} (\texttt{Numba})\\
  \texttt{celer} &\citet{Massias_Gramfort_Salmon2018}
                 & CD + working set + dual extrapolation
                 & \texttt{Python} (\texttt{Cython})\\
\texttt{cuML[cd]} &\citet{Raschka2020a}
                  & (Cyclic) Minimization along coordinates
                  & \texttt{Python}, \texttt{C++}\\
\texttt{cuML[qn]} &\citet{Raschka2020a}
                  & Orthant-Wise Limited Memory Quasi-Newton (OWL-QN)
                  & \texttt{Python}, \texttt{C++}\\
\texttt{FISTA}    &\citet{Beck_Teboulle09}
                  & ISTA + acceleration
                  & \texttt{Python}\\
  \texttt{glmnet} &\citet{Friedman_10}
                  & CD + working set + strong rules
                  & \texttt{R}, \texttt{C++}\\
  \texttt{ISTA} &\citet{Daubechies2004}
                & ISTA (Proximal GD)
                & \texttt{Python}\\
  \texttt{LARS} & \citet{Efron_Hastie_Johnstone_Tibshirani04}
                & Least-Angle Regression algorithm (LARS)
                & \texttt{Python} (\texttt{Cython})\\
  \texttt{FISTA[adaptive-1]} & \citet[Algo 4]{Liang_22}, \citet{Farrens_20}
                             & FISTA + adaptive restart
                             & \texttt{Python}\\
  \texttt{FISTA[greedy]} &\citet[Algo 5]{Liang_22}, \citet{Farrens_20}
                         & FISTA + greedy restart
                         & \texttt{Python}\\
  \texttt{noncvx-pro} & \citet{Poon_21}
                      & Bilevel optim + L-BFGS
                      & \texttt{Python} (\texttt{Cython})\\
  \texttt{skglm} &\citet{Bertrand_22}
                     & CD + working set + primal extrapolation
                     & \texttt{Python} (\texttt{Numba})\\
  \texttt{scikit-learn} &\citet{Pedregosa_11}
                     & CD
                     & \texttt{Python} (\texttt{Cython})\\
  \texttt{snapML[gpu]} &\citet{Dunner_18}
                     & CD + GPU
                     & \texttt{Python}, \texttt{C++}\\
  \texttt{snapML[cpu]} &\citet{Dunner_18}
                     & CD
                     & \texttt{Python}, \texttt{C++}\\
  \texttt{lasso.jl} & \citet{kornblith2021}
                    & CD
                    & \texttt{Julia}\\
  \bottomrule
\end{tabular}
\label{table:algo-lasso-benchmark}
\end{table}

\begin{table}[h]
  \centering
  \caption{List of datasets used in the Lasso benchmark in \autoref{sec:lasso}}
      \begin{tabular}{llS[table-format=6.0]S[table-format=7.0]S[table-format=1.1e-1,scientific-notation=true]}
        \toprule
        \textbf{Dataset} & \textbf{References} &\textbf{Samples ($\mathbf{n}$)} & {\textbf{Features} ($\mathbf{p}$)} & {\textbf{Density}} \\
        \midrule
        \emph{MEG} &\citet{mne} & 305 & 7498 & 1.0 \\        %
        \emph{news20}& \citet{news20} & 19996 & 1355191 & 0.00034  \\ %
        \emph{rcv1} &\citet{rcv1} & 20242 & 47236 &  0.0036 \\ %
        \emph{MillionSong} &\citet{Bertin-Mahieux2011} & 463715 & 90 & 1 \\ %
        \bottomrule
    \end{tabular}
    \label{table:summary_data_lasso}
\end{table}

\subsection{Support identification speed benchmark}

\begin{figure}
    \centering
    \includegraphics[width=0.8\figwidth]{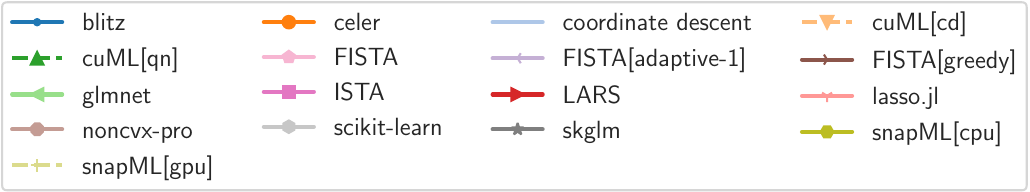}
    \includegraphics[width=\figwidth]{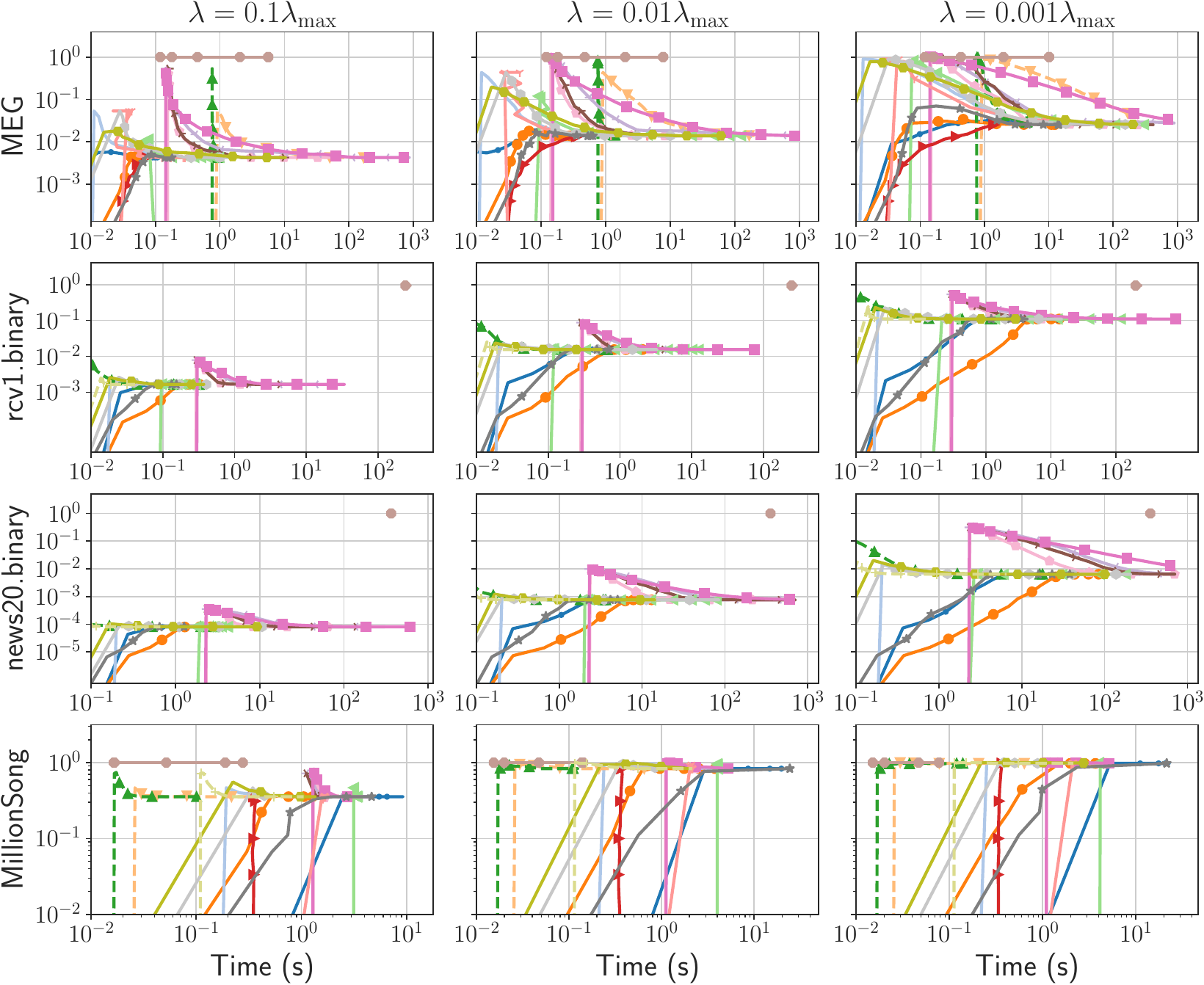}
    \caption{Additional benchmark for the Lasso on variants of the \texttt{Objective} (\emph{columns}). The curves display the fraction of non-zero coefficients in iterates $\theta_t$ ($\Vert \theta_t \Vert_0 /p$), as a function of time.}
    \label{fig:lasso_support}
\end{figure}

Since the Lasso is massively used for its feature selection properties, the speed at which the solvers identify the support of the solution is also an important performance measure.
To evaluate the behavior of solvers in this task, it is sufficient to add a single new variable in the \texttt{Objective}, namely the $\ell_0$ pseudonorm of the iterate, allowing to produce \autoref{fig:lasso_support} in addition to \autoref{fig:lasso_leukemia_meg_rcv1_news20}.

\subsection{\rebuttal{Convergence in terms of iteration}}
\autoref{app:sec:lasso_iteration}

\begin{figure}
    \centering
    \includegraphics[width=0.7\figwidth]{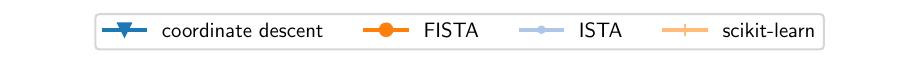}
    \includegraphics[width=\figwidth]{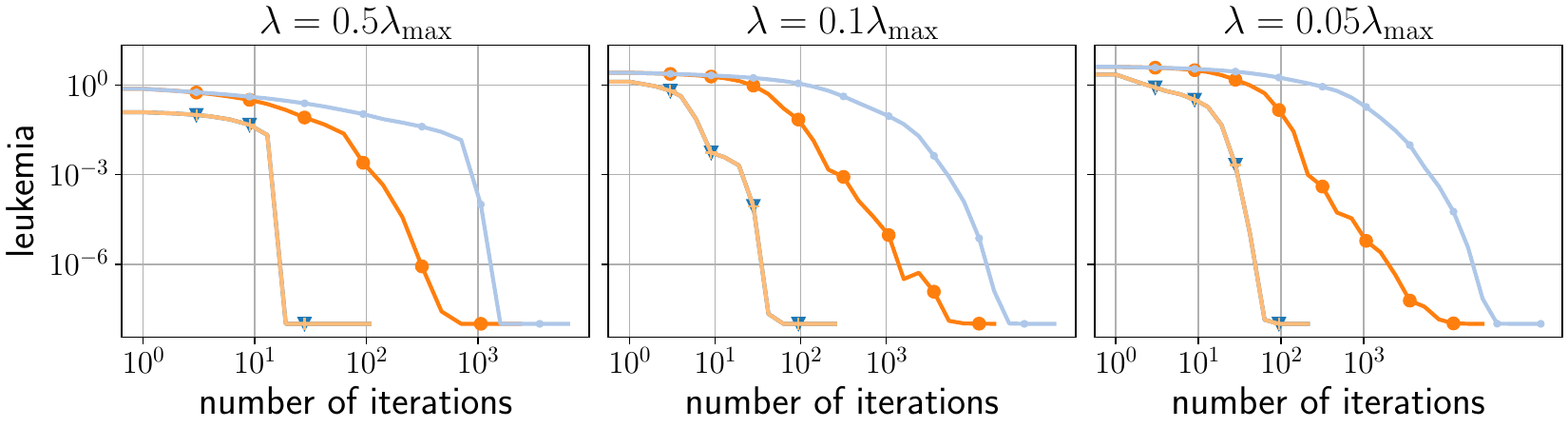}
    \caption{\rebuttal{Convergence speed with respect to the number of iterations for some solvers of the Lasso benchmark on the \emph{leukemia} dataset.}}
    \label{fig:lasso_per_iter}
\end{figure}

\rebuttal{While practitioners are mainly concerned with the time it takes to solve their optimization problem, one may also be interested in the convergence as a function of the number of iterations. This is particularly relevant to compare theoretical convergence rates with experiments. Benchopt natively supports such functionality. Yet, this makes sense only if one iteration of each algorithm costs the same. \autoref{fig:lasso_per_iter} presents such a case on the \emph{leukemia} dataset, using algorithms for which one iteration costs $n \times p$. One can observe that cyclic coordinate descent as implemented in \texttt{Cython} in \texttt{scikit-learn} or in \texttt{Numba} lead to identical results, while they outperform proximal gradient methods.}

\clearpage{}%
\clearpage{}%
\section{ResNet18}
\label{app:resnet}

\subsection{Description of the benchmark}

\myparagraph{Setting up the benchmark}
The three currently supported frameworks are \TensorFlow/\Keras~\citep{tensorflow2015-whitepaper,chollet2015keras}, \PyTorch~\citep{NEURIPS2019_9015}
and \texttt{PyTorch Lightning}~\citep{william_falcon_2020_3828935}.
We report here results for \TensorFlow/\Keras{}  and \PyTorch.
To guarantee that the model behaves consistently across the different  considered frameworks, we implemented several consistency unit tests.
We followed the best practice of each framework to make sure to achieve the optimal computational efficiency.
In particular, we tried as much as possible to use official code from the frameworks, and not third-party code.
We also optimized and profiled the data pipelines to make sure that our training was not IO-bound.
Our benchmarks were run
using \TensorFlow{} version 2.8 and \PyTorch{} version 1.10.

\myparagraph{Descriptions of the datasets}

\begin{table}[bth]
\newcolumntype{E}[1]{S[table-align-text-post=false,table-format=#1]}
\centering
\caption{Description of the datasets used in the ResNet18 image classification benchmark}
\scriptsize
\addtolength{\tabcolsep}{-2pt}
\begin{tabular}{lllcE{2.0}E{2.1}E{2.0}cc}
\toprule
\textbf{Dataset} & \textbf{Content} & \textbf{References} &{\textbf{Classes}} & {\textbf{Train Size}} & {\textbf{Val. Size}} & {\textbf{Test Size}} & {\textbf{Image Size}} & \textbf{RGB} \\
\midrule
\emph{CIFAR-10} & natural images     &   \citet{Krizhevsky09learningmultiple}   & 10 & 40k   & 10k   & 10k & 32 & \cmark \\
\emph{SVHN}    & digits in natural images & \citet{Netzer2011} & 10 & 58.6k & 14.6k & 26k & 32 & \cmark \\
\emph{MNIST}   & handwritten digits &  \citet{lecun2010mnist}     & 10 & 50k   & 10k   & 10k & 28 & \xmark \\
\bottomrule
\end{tabular}%
\medskip
\label{tab:resnet-datasets}
\end{table}

In \autoref{tab:resnet-datasets}, we present some characteristics of the different datasets used for the ResNet18 benchmark.
In particular, we specify the size of each splits when using the train-validation-test split strategy.
The test split is always fixed, and is the official one.

While the datasets are downloaded and preprocessed using the official implementations of the frameworks, we made sure to test that they matched using a unit test.

\myparagraph{ResNet}
The ResNet18 is the smallest variant of the architecture introduced by \citet{He2016}.
It consists in 3 stages:
\begin{enumerate}[topsep=0pt,itemsep=1ex,partopsep=0ex,parsep=0ex,leftmargin=3ex]
    \item A feature extension convolution that goes from 3 channels (RGB, or a repeated grayscale channel in the \emph{MNIST} case) to 64, followed by a batch normalization and a ReLU.
    \item A series of residual blocks. Residual blocks are grouped by scale, and each individual group starts with a strided convolution to reduce the image scale (except the first one). As the scale increases, so does the number of features (64, 128, 256, 512). In the ResNet18 case, each scale group has two individual residual blocks and there are four scales. A residual block is comprised of three convolution operations, all followed by a batch normalization layer, and the first two also followed by a ReLU. The input is then added to the output of the third batch normalization layer before being fed to a ReLU.
    \item A classification head that performs global average pooling, before applying a fully connected (i.e. dense) layer to obtain logits.
\end{enumerate}

\myparagraph{Training's hyperparameters}
\begin{table}[bth!]
\centering
\caption{Hyperparameters used for each solver. If a hyperparameter's value is not specified in the table, it was set as the default of the implementation (checked to be consistent across frameworks).}
\begin{tabular}{lcc}
\toprule
\textbf{Hyperparameter} & \textbf{SGD} & \textbf{Adam} \\
\midrule
Learning Rate  & 0.1               & 0.001          \\
Momentum       & 0.9               & N/A           \\
Weight Decay   & $5\times 10^{-4}$ & 0.02          \\
Batch Size     & 128               & 128           \\
\bottomrule
\end{tabular}
\medskip
\label{tab:hp-resnet}
\end{table}

In \autoref{tab:hp-resnet}, we specify the hyperparameters we used for the benchmark.
For the SGD, the values were taken from the \href{https://github.com/kuangliu/pytorch-cifar}{pytorch-cifar GitHub repository}, while for Adam we took the most relevant ones from the work of \citet{Wightman2021}.

\subsection{Hyperparameter sensitivity}
\begin{figure}[t]
    \centering
    \includegraphics[width=0.99\figwidth]{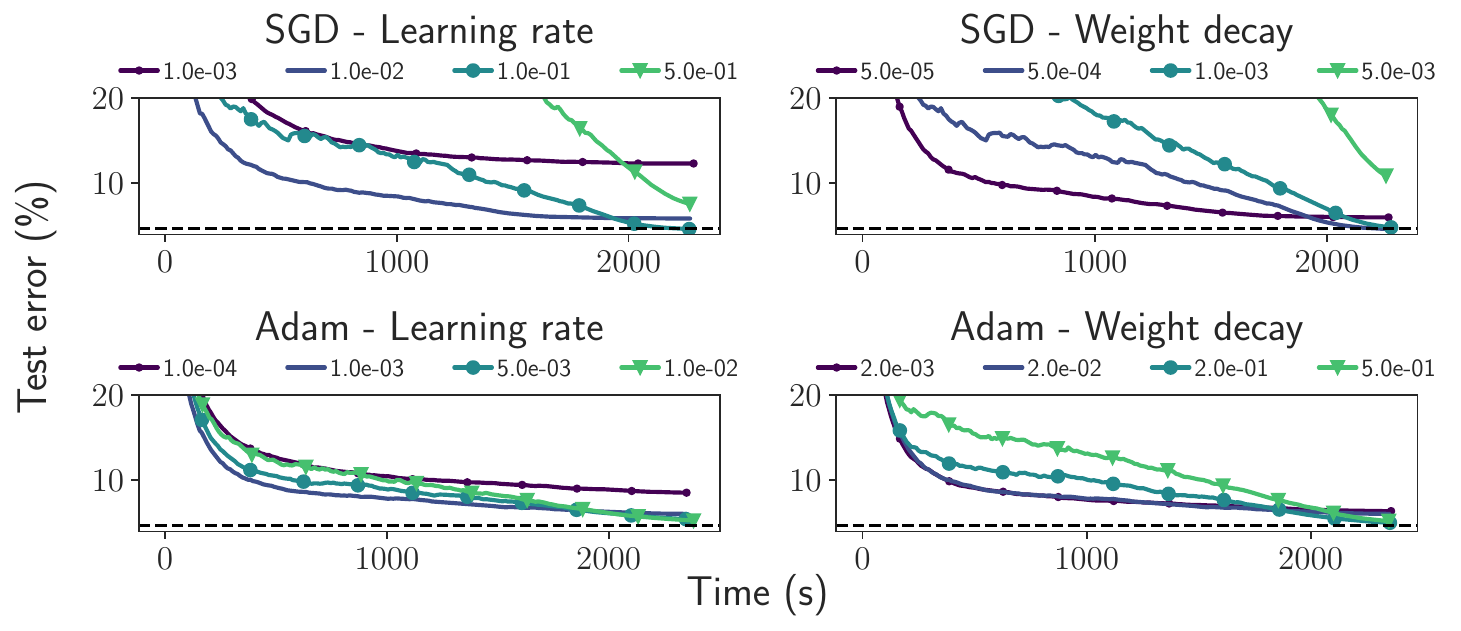}
    \caption{\textbf{ResNet18 image classification benchmark on \emph{CIFAR-10} for different values of learning rate and weight decay for SGD and Adam.} The default values are that reported in \autoref{tab:hp-resnet}. In dashed black is the state of the art for \emph{CIFAR-10} with a ResNet18 measured by \citet{zhang2019lookahead}. \repo{Curves are exponentially smoothed.}}
    \label{fig:resnet18_hp_sens}\vskip-1.1em
\end{figure}

\rebuttal{
In the benchmark presented in  \autoref{sec:resnet18}, we consider fixed hyperparameters chosen from common practices to train ResNet18 models for an image classification task.
However, in practice, these hyperparameters must be carefully set, either via a grid search, or via more adapted algorithms such as random search~\citep{bergstra2012random} or bayesian optimization~\citep{NEURIPS2019_9015}.
It is therefore important to evaluate how sensitive an optimizer is to choosing the right parameters, as  more sensitive methods will require more exhaustive hyperparameters search.
In \autoref{fig:resnet18_hp_sens}, we study this issue using \Benchopt{} for image classification on \emph{CIFAR-10}.
Despite achieving the best results in terms of accuracy, SGD is way more sensitive to the choice of hyperparameters than Adam.\footnote{We ran the same experiment on two other datasets obtaining similar figures.}
}

\rebuttal{
Another way to look at hyperparameter sensitivity is to evaluate how a given selection of hyperparameters performs for different tasks.
\autoref{fig:resnet18_sgd_torch} shows that while SGD is sensitive to the choice of learning rate and weight decay, the selected values  work very well across 3 different datasets.
}

\subsection{Aligning TensorFlow and PyTorch ResNet18 training}
\label{sec:diff-tf-pt}

\begin{table}[h!]
\centering
\caption{
Differences in off-the-shelf implementations of various components when training ResNet18 for image classification in \PyTorch{} and \TensorFlow. The selected versions are put in bold font for components that we were able to reconcile. This highlights the numerous details to consider when comparing experimental results.}
\begin{tabular}{lcc}
\toprule
\textbf{Component}                 & \textbf{\PyTorch}            & \textbf{\TensorFlow/\Keras} \\ \midrule
Bias in convolutions               & \xmark                      & \cmark                    \\
Decoupled weight decay scaling     & \textbf{Multiplied by learning rate} & Completely decoupled      \\
Batch normalization momentum       & \textbf{0.9}                & 0.99                      \\
Conv2D weights init.               & \textbf{Fan out, normal}    & Fan average, uniform      \\
Classification head init. (weights)& \textbf{Fan in, uniform}    & Fan average, uniform      \\
Classification head init. (bias)   & \textbf{Fan in, uniform}       & Zeros                  \\
Striding in convolutions           & \textbf{Starts one off}     & Ends one off              \\
Variance estimation in batch norm  & unbiased (eval)/biased (training)            & biased                    \\
\bottomrule
\end{tabular}
\label{tab:diff-tf-pt}
\end{table}

We summarized in \autoref{tab:diff-tf-pt} the different elements that have to be considered to align the training of a ResNet18 in PyTorch and TensorFlow.
Let us detail here some lines of this table:
\begin{itemize}[topsep=0pt,itemsep=1ex,partopsep=0ex,parsep=0ex,leftmargin=3ex]
    \item \textbf{Bias in convolutions:} It can be seen in \href{https://github.com/keras-team/keras/blob/master/keras/applications/resnet.py#L238}{TensorFlow/Keras official implementation}, that convolutions operations use a bias. This is in contrast to \href{https://github.com/pytorch/vision/blob/main/torchvision/models/resnet.py#L49}{PyTorch's official implementation in \texttt{torchvision}} which does not.
    Since the convolutions are followed by batch normalization layers, with a mean removal, the convolutions' bias is a spurious parameter, as was noted by \citet{ioffe2015batch}.
    We therefore chose to use unbiased convolutions.
    \item \textbf{Decoupled weight decay scaling:} this led us to scale manually the weight decay used in TensorFlow by the learning rate when setting it.
    Moreover, because the weight decay is completely decoupled from the learning rate, it is important to update it accordingly when using a learning rate schedule, as noted in \href{https://www.tensorflow.org/addons/api_docs/python/tfa/optimizers/extend_with_decoupled_weight_decay}{the TensorFlow documentation}.
    \item \textbf{Batch normalization momentum:} an important note here is that the convention used to implement the batch normalization momentum is not the same in the 2 frameworks.
    Indeed we have the relationship $\text{\texttt{momentum}}_\text{TF} = 1 - \text{\texttt{momentum}}_\text{PT}$.
    \item \textbf{Conv2D weights intialization:} TensorFlow/Keras uses the default intialization which is a Glorot uniform intialization~\citep{glorot2010understanding}.
    PyTorch uses a He normal initialization~\citep{he2015delving}.
    We used \href{https://www.tensorflow.org/api_docs/python/tf/keras/initializers/VarianceScaling}{TensorFlow's Variance Scaling framework} to differentiate the 2.
    \item \textbf{Striding in convolutions:} when using a stride of 2 in convolutions on an even-size image, one needs to specify where to start the convolution in order to know which lines (one in every two) in the image will be removed.
    The decision is different between TensorFlow and PyTorch.
    This is not expected to have an effect on the final performance, but it makes it more difficult to compare the architectures when unit testing.
    We therefore decided to align the models on this aspect as well.
    \item \textbf{Variance estimation in batch normalization:} in order to estimate the batch variance during training for batch normalization layers, it is possible to chose between the unbiased and the biased variance estimator.
    The unbiased variance estimator applies a Bessel correction to the biased variance estimator, namely a multiplication by a factor $\frac{m}{m-1}$, where $m$ is the number of samples used to estimate.
    It is to be noted that PyTorch does uses the biased estimator in training, but stores the unbiased estimator for use during inference.
    TensorFlow does not allow for such a behaviour, and the 2 are therefore not reconcilable\footnote{It is possible to use the unbiased estimator in TensorFlow for the batch normalization, even if not documented, but its application is consistent between training and inference unlike PyTorch.}.
    Arguably this inconsistency should not play a big role with large batch sizes, but can be significant for smaller batches, especially in deeper layers where the feature map size (and therefore the number of samples used to compute the estimates) is reduced.
\end{itemize}

\myparagraph{Adapting official ResNet implementations to small images} In addition to these elements, it is important to adapt the reference implementations of both frameworks to the small image case.
Indeed, for the case of ImageNet, the ResNet applies two downsampling operations (a stride-2 convolution and a max pooling) at the very beginning of the network to make the feature maps size more manageable.
In the case of smaller images, it is necessary to do without these downsampling operations (i.e. perform the convolution with stride 1 and get rid of the max pooling).

\myparagraph{Coupled weight decay in TensorFlow}
In TensorFlow, the SGD implementation does not allow the setting of coupled weight decay.
Rather, one has to rely on the equivalence (up to a scale factor of 2) between coupled weight decay and L2 regularization.
However, in TensorFlow/Keras, adding L2 regularization on an already built model (which is the case for the official ResNet implementation), is not straightforward and we relied on the workaround of \citet{silva2019kerasregularization}.

\subsection{VGG benchmark on CIFAR-10}
\label{sec:bench-vgg}
In order to show how flexible \Benchopt{} is, we also ran a smaller version of the ResNet benchmark using a VGG16~\citep{DBLP:journals/corr/SimonyanZ14a} network instead of a ResNet18.
In the \Benchopt{} framework, this amounts to specifying a different model in the objective, while all the other pieces of code in the benchmark remain unchanged.
Note that the VGG official implementations also need to be adapted to the CIFAR-10 case by changing the classification head.
This was not specified in the original paper, where no experiment was conducted on small-scale datasets, and we relied on available open source implementations (\href{https://github.com/SeHwanJoo/cifar10-vgg16}{cifar10-vgg16} and \href{https://github.com/geifmany/cifar-vgg}{cifar-vgg}) to make this decision.
Importantly, these implementations use batch normalization to make the training of VGG more robust to initialization, which is not the case in the official framework implementations.

\begin{figure}
    \centering
    \includegraphics[width=\figwidth]{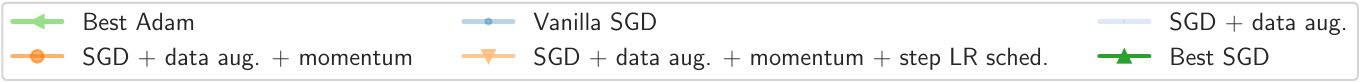}
    \includegraphics[width=0.99\figwidth]{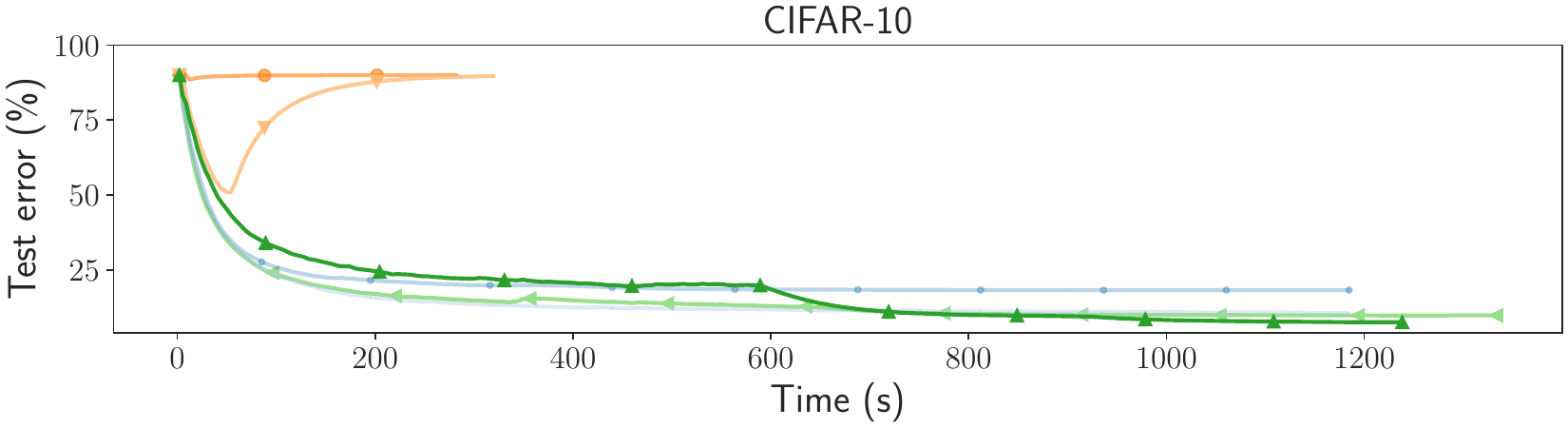}
    \caption{\textbf{VGG16 image classification benchmark with \PyTorch{} solvers.} The best SGD configuration features data augmentation, momentum, step learning rate schedule and weight decay.}
    \label{fig:bench-vgg}
\end{figure}

In \autoref{fig:bench-vgg}, we see that for the case of VGG, the application of weight decay is so important that without it, in cases with momentum, the model does not converge.\clearpage{}%
\clearpage{}%
\section{\texorpdfstring{$\ell_1$}{l1}-regularized logistic regression}
\label{app:logregl1}

This additional benchmark is dedicated to $\ell_1$-regularized logistic regression, in the same setting as \autoref{pb:logreg} but this time with an $\ell_1$-regularization for the parameters of the model:
\begin{problem}
    \label{pb:logregl1}
  \theta^*
  = \argmin_{\theta \in \bbR^p}
    \sum_{i=1}^{n} \log\big(1 + \exp(-y_i X_i^\top \theta)\big) + \lambda \|\theta\|_1
    \enspace .
\end{problem}

\subsection{List of solvers and datasets used in the $\ell_1$-regularized logistic regression benchmark}

\repo{The code for the benchmark is available at \url{https://github.com/benchopt/benchmark_logreg_l1/}.}
\autoref{table:algo-logregl1-benchmark} and \autoref{table:summary_data_logreg_l1} present the solvers and datasets used in this benchmark.
\begin{table}[h]
  \centering
  \footnotesize
  \caption{List of solvers used in the $\ell_1$-regularized logistic regression benchmark}
\begin{tabular}{p{3.2cm} p{3.3cm} p{3cm} l}
  \toprule
  \textbf{Solver} & \textbf{References} &\textbf{Description} & \textbf{Language}\\
  \midrule
  \texttt{blitz} &\citet{johnson2015blitz}
                 & CD + working set
                 & \texttt{Python}, \texttt{C++}\\
  \texttt{coordinate descent} &\citet{Friedman_10}
                              & (Cyclic) Minimization along coordinates
                              & \texttt{Python} (\texttt{Numba})\\
  \texttt{coordinate descent (Newton)} &\citet{Friedman_10}
                              & CD + Newton
                              & \texttt{Python} (\texttt{Numba})\\
  \texttt{celer} &\citet{Massias_Gramfort_Salmon2018}
                 & CD + working set + dual extrapolation
                 & \texttt{Python} (\texttt{Cython})\\
  \texttt{copt[FISTA line search]} & \citet{Pedregosa_20}, \citet{Beck_Teboulle09}
                          & FISTA (ISTA + acceleration) + line search
                          &\texttt{Python} (\texttt{Cython})\\
  \texttt{copt[PGD]} & \citet{Pedregosa_20}, \citet{Combettes2005}
                          & Proximal Gradient Descent
                          &\texttt{Python} (\texttt{Cython})\\
  \texttt{copt[PGD linesearch]} & \citet{Pedregosa_20}, \citet{ Combettes2005}
                          & Proximal Gradient Descent + linesearch
                          &\texttt{Python} (\texttt{Cython})\\
  \texttt{copt[saga]} & \citet{Pedregosa_20}
                          & SAGA (Variance reduced stochastic method)
                          &\texttt{Python} (\texttt{Cython})\\
  \texttt{copt[svrg]} &\citet{Pedregosa_20}
                           & SVRG (Variance reduced stochastic method)
                           & \texttt{Python} (\texttt{Cython})\\
  \texttt{cuML[gpu]} & \citet{Raschka2020a}
                     & L-BFGS + GPU
                     & \texttt{Python}, \texttt{C++}\\
  \texttt{cuML[qn]} & \citet{Raschka2020a}
                     & Orthant-Wise Limited Memory Quasi-Newton (OWL-QN)
                     & \texttt{Python}, \texttt{C++}\\
  \texttt{cyanure} & \citet{Mairal19}
                     & Proximal Minimization by Incremental Surrogate Optimization (MISO)
                     & \texttt{Python}, \texttt{C++} \\
  \texttt{lightning} & \citet{Blondel2016}
                         & (Cyclic) Coordinate Descent
                         & \texttt{Python} (\texttt{Cython})\\
  \texttt{scikit-learn[liblinear]} &\citet{Pedregosa_11}, \citet{Fan2008}
                                   & Truncated Newton Conjugate-Gradient
                                   & \texttt{Python} (\texttt{Cython})\\
  \texttt{scikit-learn[lbfgs]} & \citet{Pedregosa_11}, \citet{2020SciPy-NMeth}
                               & L-BFGS (Quasi-Newton Method)
                               & \texttt{Python} (\texttt{Cython})\\
  \texttt{scikit-learn[newton-cg]} &\citet{Pedregosa_11}, \citet{ 2020SciPy-NMeth}
                     & Truncated Newton Conjugate-Gradient
                     & \texttt{Python} (\texttt{Cython})\\
  \texttt{snapml[gpu=True]} & \citet{Dunner_18}
                            & CD + GPU
                            & \texttt{Python}, \texttt{C++}\\
  \texttt{snapml[gpu=False]} &\citet{Dunner_18}
                             & CD
                             & \texttt{Python}, \texttt{C++}\\
  \bottomrule
\end{tabular}
\label{table:algo-logregl1-benchmark}
\end{table}

\begin{table}[h]
  \centering
  \caption{List of the datasets used in the $\ell_1$-regularized logistic regression benchmark}
      \begin{tabular}{llS[table-format=8.0]S[table-format=7.0]S[table-format=1.1e-1,scientific-notation=true]}
        \toprule
        \textbf{Datasets} & \textbf{References} &{\textbf{Samples ($\mathbf{n}$)}} & {\textbf{Features ($\mathbf{p}$)}} & {\textbf{Density}} \\
        \midrule
        \emph{gisette} & \citet{madelon} & 6000 & 5000 & 9.9e-1 \\
        \emph{colon-cancer} & \cite{madelon} & 62 & 2000 & 1.0 \\
        \emph{news20.binary} &\citet{news20} & 19996 & 1355191 & 0.00034 \\
        \emph{rcv1.binary} &\cite{madelon} & 20242 & 19959 & 3.6e-3 \\
        \bottomrule
    \end{tabular}
    \label{table:summary_data_logreg_l1}
\end{table}

\subsection{Results}

\begin{figure}[t]
    \centering
    \includegraphics[width=.8\figwidth]{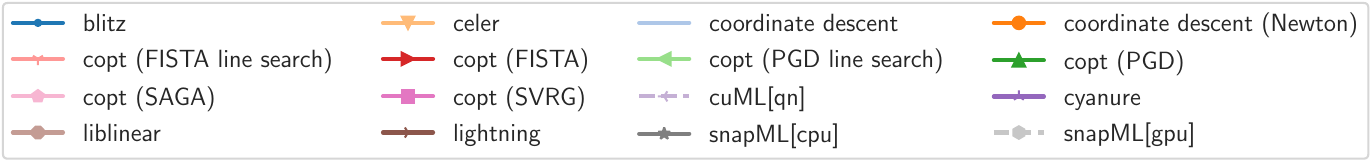}
    \includegraphics[width=0.99\figwidth]{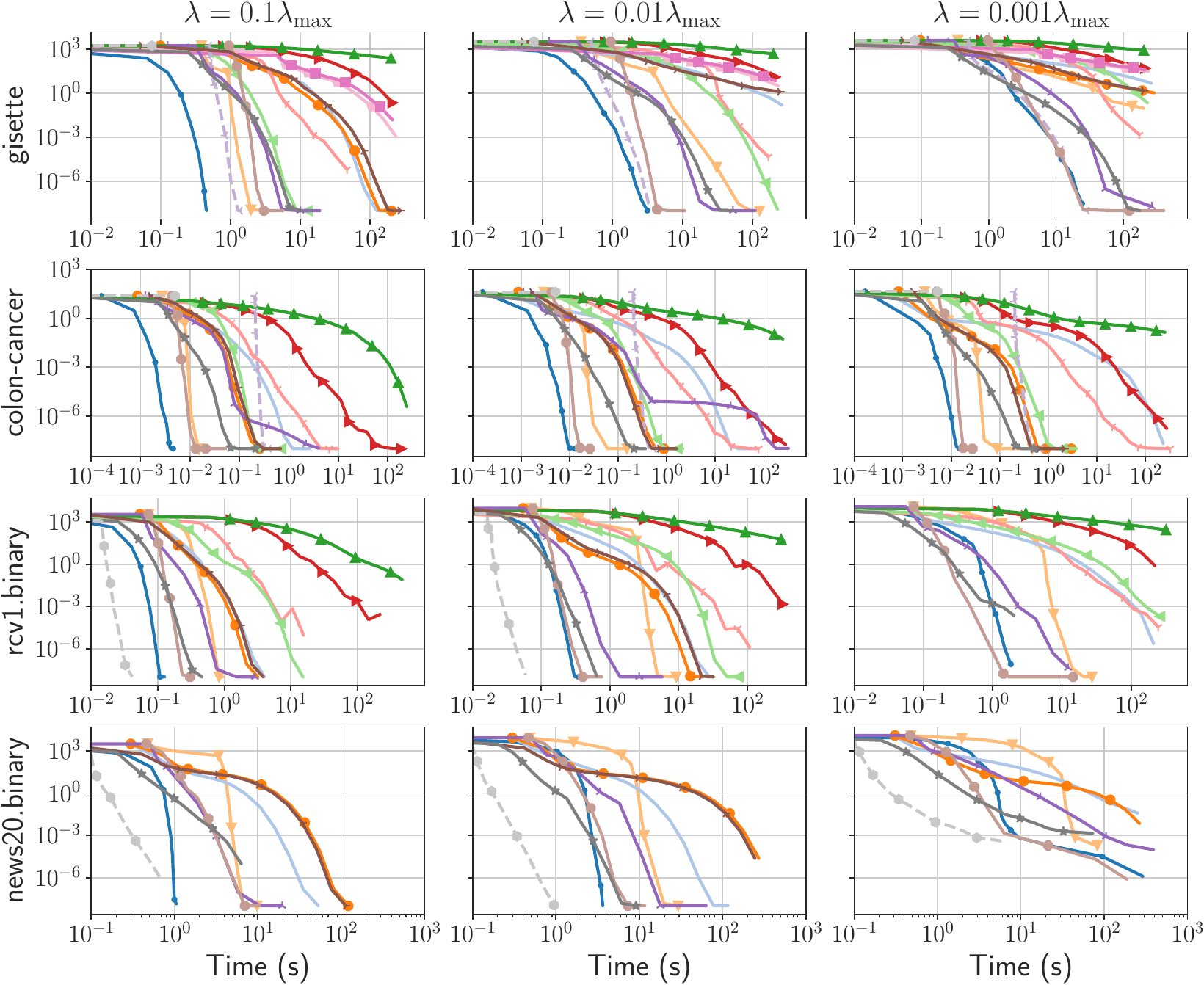}
    \caption{
        Benchmark for the $\ell_1$-regularized logistic regression on variants of the \texttt{Objective} (\emph{columns}).
        The curves display the suboptimality of the iterates, $f(\theta^t) - f(\theta^*)$, as a function of time.
        The first column corresponds to the objective detailed in \autoref{pb:logregl1} with $\lambda = 0.1  \Vert X^\top y\Vert_\infty / 2$, the second one with $\lambda = 0.01 \Vert X^\top y\Vert_\infty / 2$ and the third column with $\lambda=0.001  \Vert X^\top y\Vert_\infty / 2$.
    }
    \label{fig:logreg_l1}
\end{figure}

The results of the $\ell_1$-regularized logistic regression benchmark are in \autoref{fig:logreg_l1}.

\clearpage{}%
\clearpage{}%
\section{Unidimensional total variation}\label{sec:app:tv1d}

The use of 1D Total Variation regularization takes its root in the taut-string algorithm~\citep{barlow1972isotonic} and can be seen as a special case of either the Rudin-Osher-Fatemi model~\citep{rudin1992nonlinear} or the Generalized Lasso~\citep{tibshirani2011solution} for a quadratic data fit term.
It reads
\begin{problem}\label{pb:tv1d}
    \theta^* \in \argmin_{\theta\in \bbR^p} F(y, X \theta) + \lambda \norm{D \theta}_1 \enspace,
\end{problem}
where $F$ is a data fidelity term, $X \in \mathbb{R}^{n \times p}$ is a design matrix with $n$ samples and $p$ features, $y \in \mathbb{R}^n$ is the target vector, $\lambda > 0$ is a regularization hyperparameter, and $D \in \mathbb{R}^{(p-1) \times p}$ is a finite difference operator defined by $(D \theta)_k = \theta_{k+1} - \theta_k$ for all $1 \leq k \leq p-1$ (it is also possible to use cyclic differences).

Most often, the data fidelity term is the $\ell^2$-loss $F(y, z) = \tfrac{1}{2} \norm{y - z}_2^2$, following an additive Gaussian noise hypothesis. But the data fit term can also account for other types of noises, such as noises with heavy tails using the Huber loss $F(y, z) = | y - z |_\mu$ where $|\cdot|_\mu$ is defined coordinate-wise by
\begin{equation*}
|t|_\mu =
    \begin{cases}
    \frac{1}{2} t^2 &\textrm{if } |t| \leq \mu\\
    \mu |t|-\frac{\mu^2}{2} &\textrm{otherwise.}
    \end{cases}
\end{equation*}
\autoref{pb:tv1d} promotes piecewise-constant solutions -- alternatively said, solutions such that their gradients is sparse -- and was proved to be useful in several applications, in particular for change point detection~\citep{bleakley2011group,Tibshirani2014}, for BOLD signal deconvolution in functional MRI \citep{Karahanoglu2013,Cherkaoui2019} or for detrending in oculomotor recordings~\citep{Lalanne2020}.

The penalty $\theta \mapsto \norm{D \theta}_1$ is convex but non-smooth, and its proximity operator has no closed form. Yet as demonstrated by \citet{taut_string}, the taut-string algorithm allows to compute this proximity operator in $O(p^2)$ operations in the worst case, but it enjoys a $O(p)$ complexity in most cases.
Other methods do not rely on this proximity operator and directly solve \autoref{pb:tv1d}, using either primal-dual approaches \citep{ChambollePock2011,Condat2013}, or solving the dual problem~\citep{Pesquet2015}.
Finally, for 1-dimensional TV regularization, one can also use the synthesis formulation~\citep{SElad2007} to solve the problem.
By setting $z =  D \theta$ and $\theta = Lz + \rho$ where $L \in \mathbb{R}^{p \times p-1}$ is a lower trianglar matrix representing an integral operator (cumulative sum), the problem is equivalent to a Lasso problem, and $\rho^*$ has a closed-form expression (see \eg \citealt{bleakley2011group} for a proof).
As a consequence, any lasso solver can be used to obtain the solution of the Lasso problem $z^*$ and the solution of the original \autoref{pb:tv1d} $u^*$ is retrieved as $u^* = L z^* + \rho^*$.

\repo{The code for the benchmark is available at \url{https://github.com/benchopt/benchmark_tv_1d/} and} \autoref{table:algo-1dtv-benchmark} details the different algorithms used in this benchmark.

\begin{table}[h]
  \centering
  \small
  \caption{List of solvers used in the 1D Total Variation benchmarks}
\begin{tabular}{lllp{4.5cm}}
\toprule
  \textbf{Solver} & \textbf{References} & \textbf{Formulation} & \textbf{Description} \\
  \midrule
  \texttt{ADMM} &\citet{SBoyd2011} & Analysis
                     & Primal-Dual Augmented Lagrangian \\
  \texttt{ChambollePock} &\citet{ChambollePock2011} & Analysis
                     & Primal-Dual Hybrid Gradient \\
  \texttt{CondatVu} &\citet{Condat2013}  & Analysis
                     & Primal-Dual Hybrid Gradient \\
  \texttt{DPGD} &\citet{Pesquet2015}  & Analysis
                     & Dual proximal GD (+ acceleration)\\
  \texttt{PGD} &\citet{taut_string}  & Analysis
                     & Proximal GD + taut--string\\
               &\citet{ABarbero2018} &
                     & ProxTV (+ acceleration)\\
  \midrule\midrule
  \texttt{celer} &\citet{Massias_Gramfort_Salmon2018} & Synthesis
                     & CD + working set (lasso)\\
                     &&& \emph{only for $\ell_2$ data-fit}\\
  \texttt{FP} &\citet{Combettes2021} & Synthesis
                     & Fixed point with block updates \\
  \texttt{ISTA} &\citet{Daubechies2004} & Synthesis
                     & Proximal GD (+ acceleration)\\
  \texttt{skglm} &\citet{Bertrand_22} & Synthesis
                     & CD + working set\\
  \bottomrule
\end{tabular}
\vskip -0.1in
\label{table:algo-1dtv-benchmark}
\end{table}

\begin{figure}[t]
    \centering
    \includegraphics[width=.9\figwidth]{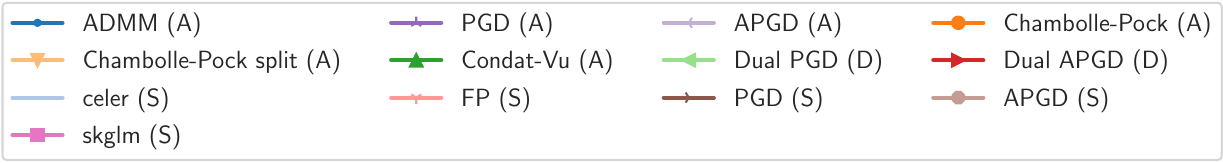}
    \includegraphics[width=0.99\figwidth]{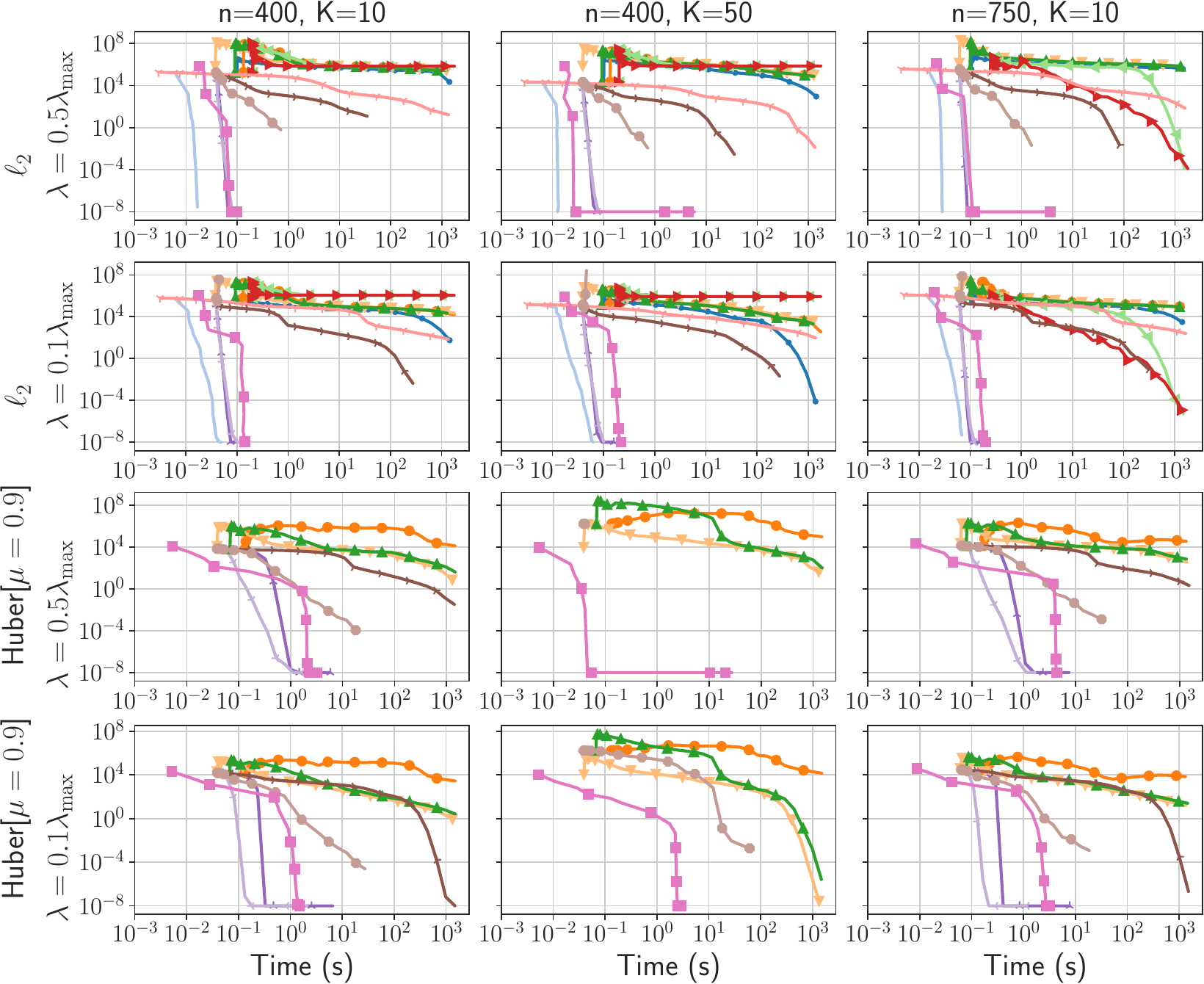}
    \caption{
        Benchmark for the $TV$-regularized regression regression, on 13 solvers, 4 variants of the \texttt{Objective} (\emph{rows}), and 3 configurations for a simulated dataset (\emph{columns}).
        The curves display the suboptimality of the iterates, $f(\theta^t) - f(\theta^*)$, as a function of time.
        The solvers in this benchmark showcase the three resolution approaches with the Analysis (A), Dual (D) and Synthesis (S) formulations.
    }
    \label{fig:tv1d}
\end{figure}

\newpage

\myparagraph{Simulated dataset}
We use here simulated data, as applications based on fMRI and EOG signals require access to open and preprocessed data that we will make available on \OpenML{}~\cite{OpenML2013} in the future.
The data are generated as follows: a block signal $\bar \theta \in \mathbb R^p$ is generated by sampling first a sparse random vector $z \in \mathbb R^p$ with $K$ non-zero coefficients positioned randomly, and taking random values following a $\mathcal{N}(0,1)$ distribution.
Finally, $\bar \theta$ is obtained by discrete integration as $\bar \theta_i = \sum_{k=1}^{i} z_k$ for $1 \leq i \leq p$.
The design matrix $X \in \mathbb R^{n \times p}$ is a Gaussian random design with  $X_{ij} \sim \mathcal N(0, 1)$.
The observations $y$ are obtained as $y = X \bar \theta + \epsilon$, with $\epsilon \sim \mathcal{N}(0,0.01)$ a Gaussian white noise.
For all experiments, we used $p=500$ and vary the number of non-zero coefficient $K$, and the number of rows $n$ of the matrix $X$.

\myparagraph{Results} \autoref{fig:tv1d} shows that the solvers using the synthesis formulation and coordinate descent-based solvers for the Lasso ($\ell_2$ data fit term) work best on this type of problem.
For the Huber data fit term, the solver using the analysis formulation and the taut-string algorithm for the proximal operator are faster.
An interesting observation from this benchmark is the behavior of the solvers based on primal-dual splitting or dual formulation. We observe that for all these solvers, the objective starts by increasing. This is probably due to a sub-optimal initialization of the dual variable compared to the primal one. While this initialization is seldom described in the literature, it seems to have a large impact on the speed of these algorithms. This shows how \Benchopt{} allows to reveal such behavior, and could lead to practical guidelines on how to select this dual initialization.

\myparagraph{Extensions} We plan to extend this benchmark in the future to consider higher dimensional problems -- \eg{} 2D TV problems for images -- or higher order TV regularization, such as Total Generalized Variation~\cite{bredies2010total} or inf-convolution of TV functionals~\cite{chambolle1997image} -- used of instance for change point detection \citep{Tibshirani2014}. Yet, for 2D or higher dimensional problems, we can no longer use the synthesis formulation. It is however possible to apply the taut-string method of \citet{taut_string} and graph-cut methods of \citet{Boykov2001} and \citet{Kolmogorov2004} for anisotropic TV, and dual or primal-dual methods for isotropic, such as Primal-Dual Hybrid Gradient algorithm~\citep{ChambollePock2011}.
\clearpage{}%
\clearpage{}%
\section{Linear regression with minimax concave penalty (MCP)}\label{sec:app:mcp}

The Lasso problem \citep{Tibshirani96} is a least-squares regression problem with a convex non-smooth penalty that induces sparsity in its solution.
However, despite its success and large adoption by the machine learning and signal processing communities, it is plagued with some statistical drawbacks, such as bias for large coefficients.
To overcome these issues, the standard approach is to consider non-convex sparsity-inducing penalties.
Several penalties have been
proposed: \emph{Smoothly Clipped Absolute Deviation} (SCAD, \citealt{fan2001variable}), the \emph{Log Sum penalty}  \citep{candes2008enhancing}, the \emph{capped-$\ell_1$ penalty} \citep{zhang2010analysis} or the \emph{Minimax Concave Penalty} (MCP, \citealt{zhang2010nearly}).

This benchmark is devoted to least-squares regression with the latter, namely the problem:
\begin{problem}\label{pb:mcp}
    \theta^* \in \argmin_{\theta\in \bbR^p} \tfrac{1}{2 n} \norm{y - X \theta}^2 + \sum_{j=1}^p \rho_{\lambda, \gamma}(\theta_j)
    \enspace,
\end{problem}
where $X \in \mathbb{R}^{n \times p}$ is a design matrix containing $p$ features as columns, $y \in \mathbb{R}^n$ is the target vector, and $\rho_{\lambda,\gamma}$ the penalty function that reads as:
$$
\rho_{\lambda,\gamma}(t) =
\begin{cases}
     \lambda |t| - \frac{t^2}{2\gamma} \enspace, \quad &\text{if } |t| \leq \gamma  \lambda  \enspace,  \\
     \frac{\lambda^2 \gamma}{2} \enspace, \quad  &\text{if } |t| > \gamma  \lambda \enspace.
\end{cases}
$$

Similarly to the Lasso, \autoref{pb:mcp} promotes sparse solutions but the optimization problem raises some difficulties due to the non-convexity and non-smoothness of the penalty. Nonetheless, several efficient algorithms have been derived for solving it. The ones we use in the benchmark are listed in
\autoref{table:algo-mcp-benchmark}.

\begin{table}[h]
  \centering
  \small
  \caption{List of solvers used in the MCP benchmark}
\begin{tabular}{l p{5cm} l}
\toprule
  \textbf{Solver} & \textbf{References} & \textbf{Short Description} \\
  \midrule
  \texttt{CD} &\citet{breheny2011coordinate}, \citet{mazumder2011sparsenet}
                     & Proximal coordinate descent \\
   \texttt{PGD} &\citet{bolte2014proximal} &  Proximal gradient descent \\
 \texttt{GIST} & \citet{gong2013general} & Proximal gradient + Barzilai-Borwein rule \\
 \texttt{WorkSet CD} & \citet{boisbunon2014active}  & Coordinate descent + working set \\
 \texttt{skglm} & \citet{Bertrand_22} & Accelerated coordinate descent + Working set \\
  \bottomrule
\end{tabular}
\vskip -0.1in
\label{table:algo-mcp-benchmark}
\end{table}

\repo{The code for the benchmark is available at \url{https://github.com/benchopt/benchmark_mcp/}.}
For this benchmark, we run the solvers on the \emph{colon-cancer} dataset and on the simulated dataset described in \autoref{app:simulated_dataset}.
We use a signal-to-noise ratio $\texttt{snr}=3$, a correlation $\rho=0.6$ with $n=500$ observations and $p=2000$ features.

\subsection{Results}
\begin{figure}[t]
    \centering
    \includegraphics[width=.6\figwidth]{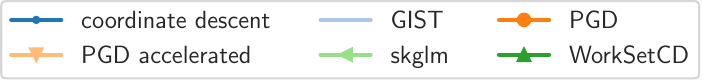}
    \includegraphics[width=0.99\figwidth]{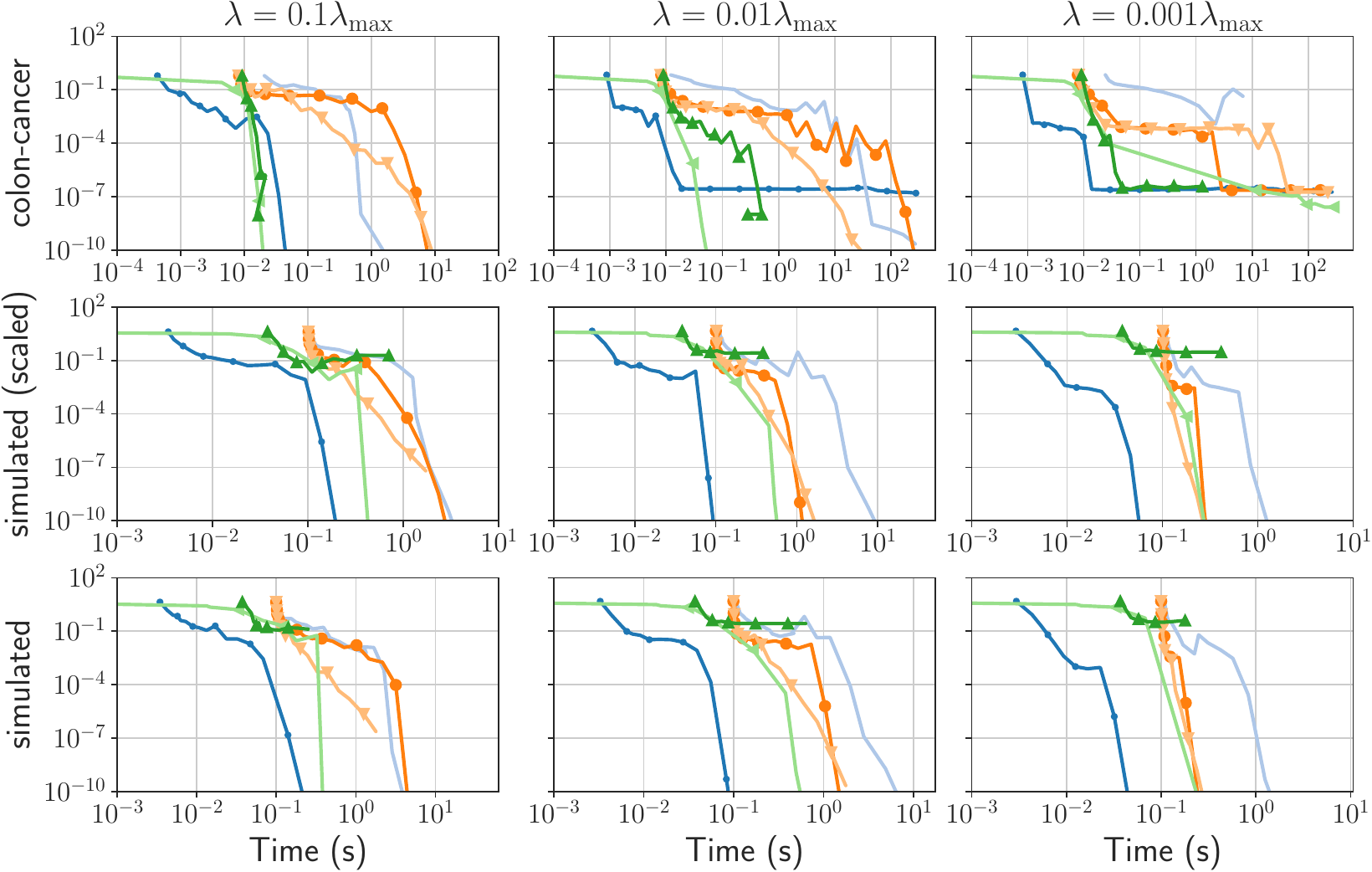}
    \caption{
        Benchmark for the MCP regression on variants of the \texttt{Objective} (\emph{columns}).
        The curves display the violation of optimality conditions, $\dist(- X^\top (X \theta_t - y) / n, \partial \rho_{\lambda, \gamma}(\theta_t))$, as a function of time.
        $\gamma$ is set to 3, and $\lambda$ is parameterized as a fraction of the Lasso's $\lambda_{\max}$, $\norm{X^\top y}_\infty / n$.
    }
    \label{fig:mcp}
\end{figure}

The result of the benchmark is presented in \autoref{fig:mcp}
The problem is non-convex and solvers are only guaranteed to converge to local minima; hence in \autoref{fig:mcp} we monitor the distance of the negative gradient to the Fréchet subdifferential of the MCP, representing the violation of the first order optimality condition.
Other metrics, such as objective of iterates sparsity, are monitored in the full benchmark, allowing to compare the different limit points obtained by the solvers.

\clearpage{}%
\clearpage{}%
\section{\rebuttal{Zero-order optimization on standard functions}}\label{sec:app:zero-order}

\rebuttal{Zero-order optimization refers to scenarios where only calls to the function to minimize are possible. This is in contrast with first-order optimization where gradient information is available. Grid search, random search, evolution strategies (ES) or Bayesian optimization (BO) are popular methods to tackle such a problem and are most commonly employed for hyperparameter optimization. This setting is also known as black-box optimization.}

\rebuttal{This benchmark demonstrates the usability of \Benchopt{} for zero-order optimization considering functions classically used in the literature~\citep{hansen:hal-01294124}. The functions are available in the PyBenchFCN package \url{https://github.com/Y1fanHE/PyBenchFCN/}. In particular, among the 61 functions of interest we present here (see \autoref{fig:zero-order}) a benchmark for three functions, defined for  any $\mathbf {x} =(x_{1},\ldots ,x_{N})\in \mathbb {R} ^{N}$:
\begin{align*}
    f(\mathbf {x} ) & = \sum _{i=1}^{N-1}[100(x_{i+1}-x_{i}^{2})^{2}+(1-x_{i})^{2}] \, &  \text{(Rosenbrock)} \\
    f(\mathbf {x} ) & = 10 \cdot N + \sum _{i=1}^{N}[(x_{i}^2-10\cdot \cos(2\pi x_i)]
    \, &\text{(Rastrigin)} \\
    f(\mathbf {x} ) & = -20 \cdot \exp \left[-0.2 {\sqrt {\frac{1}{d}\sum_{i=1}^N x_i^2}}\right] -\exp \left[  {\frac{1}{d}\sum_{i=1}^N \cos(2\pi x_i)}\right]+e+20\, & \text{(Ackley)}
    \enspace.
\end{align*}
For each function, the domain is restricted to a box: $\|\mathbf{x}\|_\infty \leq 32$ for Ackley, $\|\mathbf{x}\|_\infty \leq 30$ for Rosenbrock and $\|\mathbf{x}\|_\infty \leq 5.12$ for Rastrigin.
The algorithms considered in the benchmark are listed in
\autoref{table:algo-zero-order-benchmark}. As \texttt{BFGS} requires first-order information, gradients are approximated with finite-differences.}

\begin{table}[h]
  \centering
  \small
  \caption{List of solvers used in the zero-order benchmark}
\begin{tabular}{l p{3.5cm} l}
\toprule
  \textbf{Solver} & \textbf{References} & \textbf{Description} \\
  \midrule
   \texttt{Basin-hopping} &\citet{wales1997global,2020SciPy-NMeth} &  Two-phase method:  global step + local min. \\
\texttt{Nevergrad-RandomSearch} & \citet{nevergrad,Bergstra_Bengio12} & Sampler by random search\\
\texttt{Nevergrad-CMA} & \citet{nevergrad,hansen2001completely} & CMA evolutionary strategy \\
\texttt{Nevergrad-TwoPointsDE} & \citet{nevergrad} & Evolutionary strategy \\
\texttt{Nevergrad-NGOpt} & \citet{nevergrad} & Adaptive evolutionary algorithm \\
\texttt{Nelder-Mead} & \citet{gao2012implementing,2020SciPy-NMeth} & Direct search (downhill simplex)\\
\texttt{BFGS} & \citet{2020SciPy-NMeth} & BFGS with finite differences\\
\texttt{Powell} & \citet{powell1964efficient,2020SciPy-NMeth} & Conjugate direction method\\
\texttt{optuna-TPE} & \citet{optuna,bergstra13} & Sampler by Tree Parzen Estimation (TPE)\\
\texttt{optuna-CMA} & \citet{optuna,hansen2001completely} & CMA evolutionary strategy \\
\bottomrule
\end{tabular}
\vskip -0.1in
\label{table:algo-zero-order-benchmark}
\end{table}

\rebuttal{\repo{The code for the benchmark is available at \url{https://github.com/benchopt/benchmark_zero_order/}.}}

\subsection{\rebuttal{Results}}
\begin{figure}[t]
    \centering
    \includegraphics[width=1\figwidth]{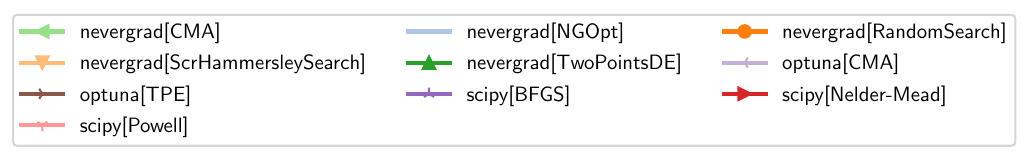}
    \includegraphics[width=0.9\figwidth]{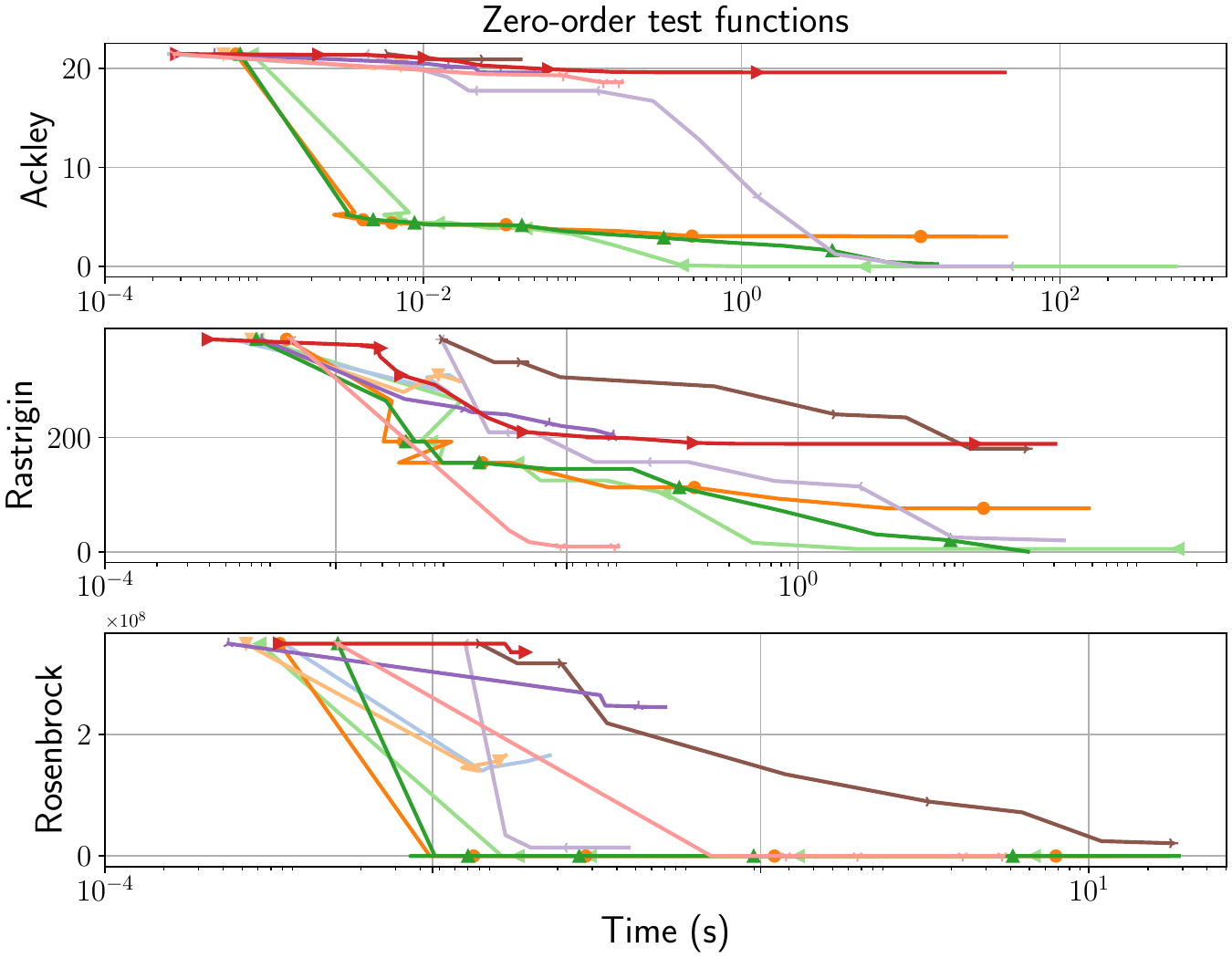}
    \caption{
        \rebuttal{
        Benchmark for the zero-order optimization on the Ackley, Rosenbrock
        and Rastrigin functions in dimension $N=20$.
        }
    }
    \label{fig:zero-order}
\end{figure}

\rebuttal{The results of the benchmark are presented in \autoref{fig:zero-order}.
The functions are non-convex and solvers are only guaranteed to converge to local minima; hence in \autoref{fig:zero-order} we monitor the value of the function. The functions are designed such that the global minimum of the function is always 0. One can observe that the CMA and TwoPointsDE implementations from \texttt{nevergrad} consistently reaches the global minimum.
In addition, the CMA implementation from \texttt{optuna} is a bit slower than the one from \texttt{nevergrad}. Also one can notice that random search offers reasonable results.
The TPE method seems to suffer from the curse of dimensionality, as most kernel methods in non-parametric estimation. Finally regarding the \texttt{scipy} solvers, \texttt{Powell} can be competitive, while \texttt{Nelder-Mead} and \texttt{BFGS} suffer a lot from local minima.}

\clearpage{}%

\end{document}